\documentclass[10pt,a4paper,twoside,american]{article}
\usepackage{geometry}
\usepackage[toc,page,titletoc]{appendix}
\usepackage{authblk}
\usepackage{calc}
\usepackage{graphicx}
\usepackage[table]{xcolor}
\definecolor{shadecolor}{gray}{0.9}
\usepackage{natbib} 
\usepackage[font=small,labelfont=bf]{caption}
\usepackage{amsmath}
\usepackage{amssymb}
\usepackage{mathtools}
\usepackage{bm}
\usepackage[font=small,labelfont=bf]{caption}
\usepackage{subcaption}
\usepackage{algorithmic}
\usepackage{lineno}

\usepackage[
breaklinks=true,
colorlinks=true,
linkcolor=gray,citecolor=gray,urlcolor=gray,filecolor=gray,
pdffitwindow,
bookmarks=true,
bookmarksopen=true,
bookmarksnumbered=true,
pdftitle={Sequence learning on a memristive array},
pdfauthor={XXX},
pdfsubject={},
pdfkeywords={}
]{hyperref}

\usepackage{cleveref}
\crefname{supp}{supplement}{Supplements}
\crefname{app}{appendix}{Appendices}
\crefformat{equation}{Eq~(#2#1#3)}
\crefformat{figure}{Fig.\,#2#1#3}
\crefformat{section}{Sec.\,#2#1#3}
\crefformat{appendix}{App.\,#2#1#3}
\crefformat{table}{Tab.\,#2#1#3}

\usepackage[final]{changes}
\definecolor{YB}{RGB}{0,150,255}
\definecolor{TT}{RGB}{0,200,0}
\definecolor{DW}{RGB}{200,0,200}
\definecolor{SIS}{RGB}{100,0,100}
\definechangesauthor[color=YB]{YB}
\definechangesauthor[color=TT]{TT}
\definechangesauthor[color=DW]{DW}
\definechangesauthor[color=SIS]{SIS}

\newcommand{\panellabel}[1]{{\bf{}#1}}  
\newcommand{\panel}[1]{{\bf{}\panellabel{#1})}}  
\newcommand{\seq}[1]{\ensuremath{\{\text{#1}\}}}

\newcommand*{\ie}{i.e.}

\newcommand{\inh}{\textnormal{I}}     
\newcommand{\exc}{\textnormal{E}}     
\newcommand{\external}{\textnormal{X}}     
\newcommand{\CM}{C_\textnormal{m}}    

\newcommand{\dtsim}{\Delta t}

\newcommand{\EE}{{\exc\exc}}
\newcommand{\EI}{{\exc\inh}}

\newcommand{\EX}{{\exc\external}}    

\newcommand{\IE}{{\inh\exc}}
\newcommand{\II}{{\inh\inh}}
\newcommand{\Epop}{\mathcal{E}} 
\newcommand{\Ipop}{\mathcal{I}} 

\newcommand{\G}{G}
\newcommand{\Xpop}{\mathcal{X}}

\newcommand{\GEI}{\G_{\exc\inh}}

\newcommand{\GIE}{\G_{\inh\exc}}

\newcommand{\GEX}{\G_\textnormal{EX}}

\newcommand{\KEE}{{K_{\exc\exc}}}
\newcommand{\KEI}{{K_{\exc\inh}}}

\newcommand{\KIE}{{K_{\inh\exc}}}
\newcommand{\KII}{{K_{\inh\inh}}}

\newcommand{\ms}{\,\textnormal{ms}}
\newcommand{\nm}{\,\textnormal{nm}}

\newcommand{\mV}{\,\textnormal{mV}}
\newcommand{\muS}{\,\textnormal{$\mu{}$S}}
\newcommand{\muA}{\,\textnormal{$\mu{}$A}}

\newcommand{\NE}{{N_\exc}}
\newcommand{\NI}{{N_\inh}}
\newcommand{\nE}{{n_\exc}}
\newcommand{\nI}{{n_\inh}}

\newcommand{\Vreset}{V_\textnormal{r}}

\newcommand\sfl{10}         
\newcommand\sfi{20}        
\newcommand\sfh{30}        

\newcommand\lplusb{0.04}       
\newcommand\lminusb{0.01}      
\newcommand\lplusa{0.1}        
\newcommand\lminusa{0.03}      
\newcommand\mplus{0.5}         
\newcommand\mminus{0.5}        
\newcommand\sw{0.01}           
\newcommand\sr{0.03}           

\title{Sequence learning in a spiking neuronal network with memristive synapses}

\def\shorttitle{Spiking memristive TM}

\author[1,2,3,*]{Younes Bouhadjar}
\author[2,3]{Sebastian Siegel}
\author[1]{Tom Tetzlaff}
\author[1,4]{Markus Diesmann}
\author[2,5]{Rainer Waser}
\author[5]{Dirk J.~Wouters}

\affil[1]{\footnotesize%
  Institute of Neuroscience and Medicine (INM-6), \& Institute for Advanced Simulation (IAS-6), \& JARA BRAIN Institute Structure-Function Relationships (INM-10), J\"ulich Research Centre, J\"ulich, Germany}
\affil[2]{\footnotesize%
  Peter Gr\"unberg Institute (PGI-7,10), J\"ulich Research Centre and JARA, J\"ulich, Germany}
\affil[3]{\footnotesize%
  RWTH Aachen University, Aachen, Germany}
\affil[4]{\footnotesize%
  Department of Physics, Faculty 1, \& Department of Psychiatry, Psychotherapy, and Psychosomatics, Medical School, RWTH Aachen University, Aachen, Germany}
\affil[5]{\footnotesize%
  Institute of Electronic Materials (IWE 2) \& JARA-FIT, RWTH Aachen University, Aachen, Germany}

\affil[*]{\footnotesize\url{y.bouhadjar{at}fz-juelich.de}}

\date{\footnotesize\today}

\usepackage{fancyhdr}
\begin{document}

\maketitle

\pagestyle{fancy}

\maketitle


\begin{abstract}

Brain-inspired computing proposes a set of algorithmic principles that hold promise for advancing artificial intelligence. 
They endow systems with self learning capabilities, efficient energy usage, and high storage capacity.
A core concept that lies at the heart of brain computation is sequence learning and prediction. 
This form of computation is essential for almost all our daily tasks such as movement generation, perception, and language.
Understanding how the brain performs such a computation is not only important to advance neuroscience but also to pave the way to new technological brain-inspired applications.
A previously developed spiking neural network implementation of sequence prediction and recall learns complex, high-order sequences in an unsupervised manner by local, biologically inspired plasticity rules. 
An emerging type of hardware that holds promise for efficiently running this type of algorithm is neuromorphic hardware.
It emulates the way the brain processes information and maps neurons and synapses directly into a physical substrate. 
Memristive devices have been identified as potential synaptic elements in neuromorphic hardware.
In particular, redox-induced resistive random access memories (ReRAM) devices stand out at many aspects.
They permit scalability, are energy efficient and fast, and can implement biological plasticity rules.
In this work, we study the feasibility of using ReRAM devices as a replacement of the biological synapses in the sequence learning model.
We implement and simulate the model including the ReRAM plasticity using the neural simulator NEST.
We investigate two types of ReRAM devices: (i) an analog switching memristive device, where the conductance gradually changes between a low conductance (LCS) and a high conductance state (HCS), and (ii) a binary switch memristive device, where the conductance abruptly changes between the LCS and the HCS.
We study the effect of different device properties on the performance characteristics of the sequence learning model, and demonstrate resilience with respect to different on-off ratios, conductance resolutions, device variability, and synaptic failure.

\end{abstract}



\section{Introduction}

In everyday's tasks such as learning, recognizing, or predicting objects in a noisy environment, the brain
outperforms conventional computing systems and deep learning algorithms at many aspects: it has a higher capacity to generalize, can learn from small training examples, is robust with respect to perturbations and failure, and is highly resource and energy efficient.
To achieve this performance, it uses intricate biological mechanisms and principles.
Understanding these principles is essential for driving new advances in neuroscience and for developing new real-world applications.
For instance, it is known that biological neural networks are highly sparse in activity and connectivity and they can self-organize in the face of the incoming sensory stimulus using unsupervised local learning rules.
A number of biologically inspired algorithms relying on these principles have been developed for sequence prediction and replay \citep{Lazar09, Hawkins16_23, Bouhadjar22_e1010233}, pattern recognition \citep{Masquelier07_e31, Payeur21_1780}, and decision making \citep{Neftci19_133}.
The spiking temporal memory (spiking TM) network proposed by \citet{Bouhadjar22_e1010233} learns high-order sequences in an unsupervised, continuous manner using local learning rules. 
Owing to its highly sparse activity and connectivity, it provides an energy-efficient sequence learning and prediction mechanism.
\par
The spiking TM algorithm was originally implemented using the neural simulator NEST \citep{Gewaltig_07_11204}.
While NEST provides a simulation platform optimized for running large-scale networks efficiently in a reproducible manner, it is executed on standard von-Neumann-type computers, i.e., on hardware that is not specifically optimized for neuromorphic computing. 
This results in performance limitations as the simulation time and the energy dissipation become substantially high for brain-scale neural networks \citep{Kunkel14_78, Jordan18_2}. 
For using spiking TM in edge-computing applications, more efficient hardware is therefore required.
Neuromorphic hardware, with
dedicated solutions to the high demands imposed by the natural-density connectivity of the brain and the resulting communication load, as well as, specific circuit blocks emulating neuron and synapse functionalities, present a potential solution for that \citep{Burr16_89, Xia19_309, Markovi20_499, Zhu20_011312}.
The local learning rules and the sparse neuronal activation of the spiking TM model allow for efficient mapping of the algorithm on neuromorphic hardware.
\par
\begin{figure}[!h]
  \centering
  \includegraphics{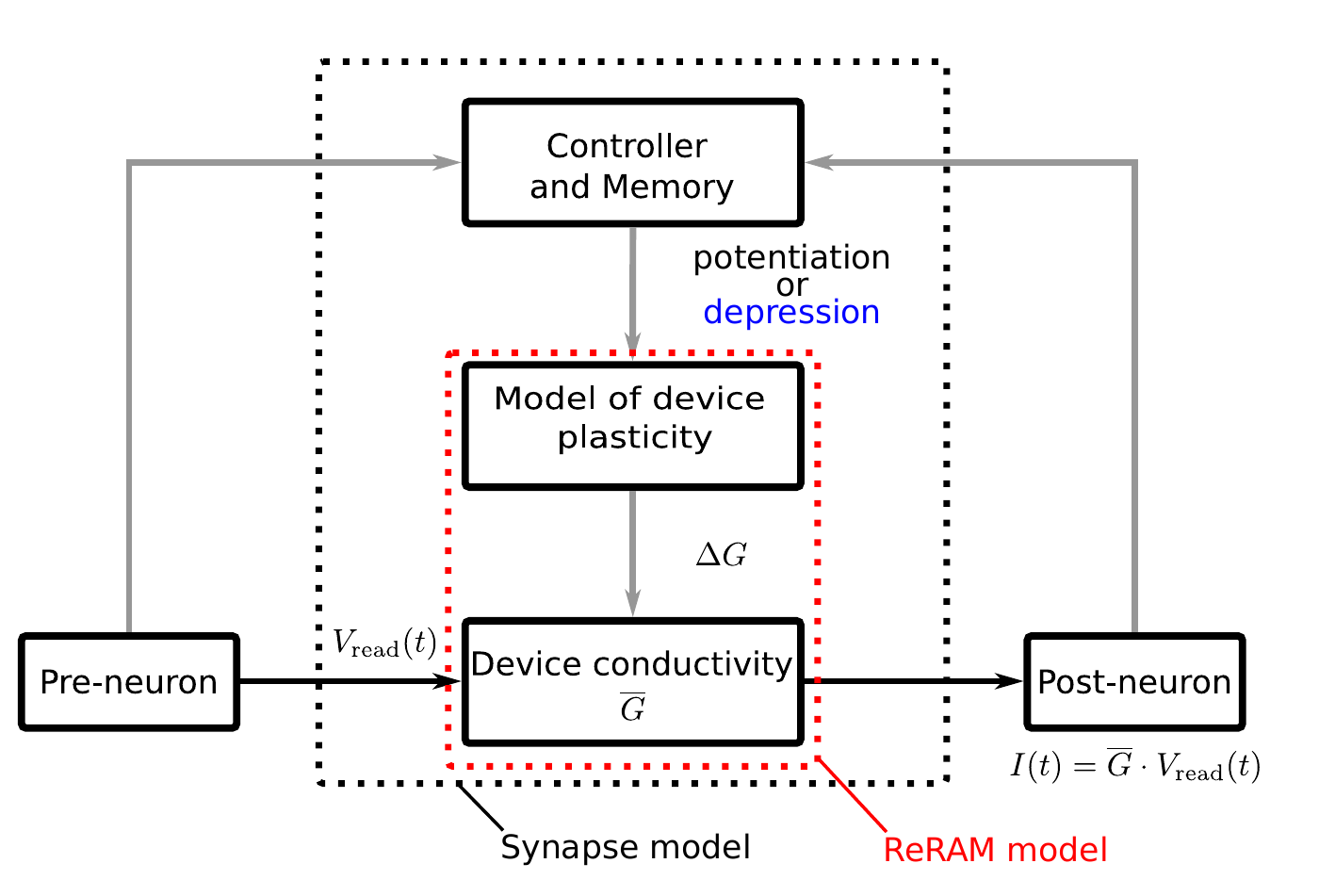}
  \caption{%
    \textbf{ReRAM control circuit.}
    Sketch depicting the synapse model including the control circuit and the ReRAM model (red box).
    The circuit is composed of a read/inference path (black arrows) and a write/programming path (gray arrows).
    The device conductivity $\overline{G}$ is read out whenever a presynaptic neuron emits a spike, which results in a postsynaptic current $I(t)=\overline{G}\cdot V_\text{read}$.
    The device conductivity is updated by the programming path.
    The controller receives pre- and postsynaptic spikes and decides on applying either a depression or a potentiation event (or both).
    In the next step, the model of device plasticity computes the conductance increment/decrement $\Delta{}G$.
  }
\label{fig:controller_memristor} 
\end{figure}

Memristive devices were suggested as components in such a hardware \citep{Yang13_13, Ielmini18_333, Yu18_260}. 
They can be used to emulate certain synaptic functionalities using only a single device, by replacing more complex CMOS-based circuits \citep{Waser09_2632, Dittmann19_110903}.
Their intrinsic dynamics capture similar characteristics as the biological synapses such as variability, weight dependence of the update, and nonvolatility.
However, while single memristive devices may readily emulate the inference function, they cannot emulate on their own plasticity rules such as spike-timing-dependent plasticity (STDP) or homeostatic control. 
The change of the memristive conductivity depends on the momentary voltage difference between its two terminals, and the device has no memory of past spike events at either of its terminals nor of their relative timing.
%
%
Hebbian learning such as STDP therefore can only be emulated using a memristive device by ``reshaping'' of the pre- and post-synaptic spike events using complex voltage pulses, so that the spike-time dependency is translated into a desired instantaneous voltage difference over the device \citep{ZamarreoRamos11_26, Wang15_438}. 
As a result, the learning rule is controlled outside the actual device (see \cref{fig:controller_memristor}).
As for implementing the learning, instead of using complex voltage pulse shapes, it is more efficient to use a controller to generate simple rectangular voltage pulses that can effectuate the desired change of the device conductance in a better, more energy efficient, and also more reliable way. 
The change of the device conductivity as a function of the number of applied voltage pulses can hereby be seen as an intrinsic plasticity curve of the device, where the actual pulse shape can be optimized towards desired potentiation and depression characteristics.
\par
In this work, we investigate how the intrinsic potentiation and depression characteristics of memristive devices influence the learning of the model in \citep{Bouhadjar22_e1010233}. 
Thereto, we adapt the original neuroscientific synapse model to accommodate memristive-type potentiation/depression characteristics.
The performance of the system is assessed by varying device characteristics such as conductance values and ranges, granularity of conductance change, and device variability. 
In this work, we study a particular type of memristive device known as the valence change memory (VCM) ReRAM device \citep{Waser12_7628}.
We investigate its two operation modes \citep{Cueppers19_091105}: either the continuous, analog mode, where the conductivity changes gradually between a Low conductance state (LCS) and a high conductance state (HCS), or the binary mode, where the conductivity changes abruptly between the LCS and the HCS. 
The binary switching is controlled by the value of an analog adaptable internal state variable \citep{Doevenspeck18_20, Zhao19, Suri13_2402, Yu18_260}. It resembles the learning rule employed in \citep{Bouhadjar22_e1010233} and mimics a structural form of STDP known in the neuroscientific literature.

\section{Results}

\subsection{A synapse model for ReRAM devices}
\label{sec:reram_model}

In this section, we briefly review the ReRAM device dynamics, introduce our model of the ReRAM device and its control circuitry, and characterize the resulting model dynamics.
\par
\begin{figure}[!h]
  \centering
  \includegraphics[width=0.25\linewidth]{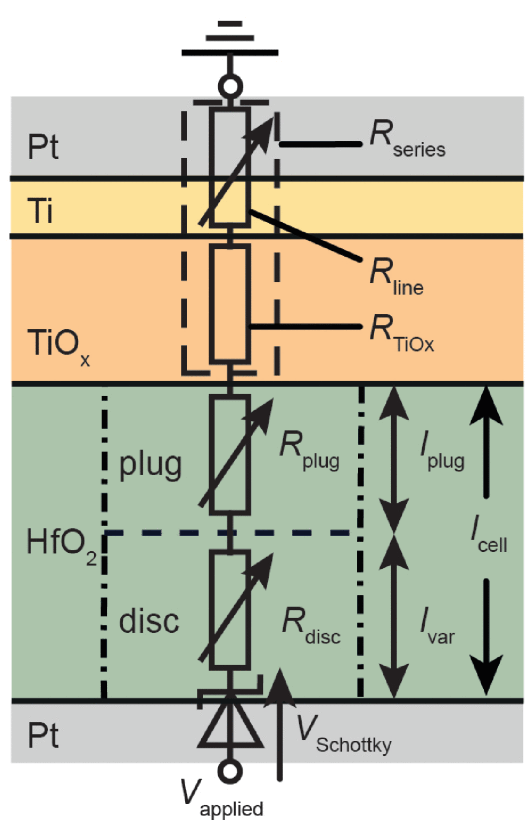}
  \caption{%
    \textbf{ReRAM stack.}
    Equivalent circuit diagram for JART VCM model describing the Pt/HfO2/TiOx/Pt (HOTO) device.
    Figure by \cite{Bengel20_4618} licensed under a Creative Commons Attribution 4.0 International License (\url{http://creativecommons.org/licenses/by/4.0/}).
  }
\label{fig:reram_stack} 
\end{figure}

The VCM ReRAM device consists of a thin ($3$--$10\nm$) insulating transition metal-oxide film sandwiched between a low work function (WF), ohmic electrode, and a high WF, blocking Schottky-interface electrode (see \cref{fig:reram_stack}).
In a first step, a conductive filament 
is formed due to a high voltage applied across the device.
This filament consists of oxygen vacancies that act as local dopant elements in the insulating metal-oxide matrix. 
During further operation, these oxygen vacancies can be moved by means of a high electric field and a local Joule heating (as a function of the polarity, either towards or away from the blocking electrode). 
The device conductance is controlled by the concentration of oxygen vacancies $N_\text{VO}$ in a small region (gap) near the electrical blocking electrode: at low $N_\text{VO}$, the filament is ``broken'', which gives rise to a high conduction barrier between the high WF electrode and the remaining filament (plug), \ie, the device is in the LCS.
If $N_\text{VO}$ is high, we have a ``connecting'' filament, where the high $N_\text{VO}$ lowers the conduction potential barrier at the blocking electrode, \ie, the device is in the HCS. 
Hence, the concentration of the oxygen vacancies $N_\text{VO}$ can be seen as an internal state variable of the device.
\par
As mentioned above, depending on the initial resistance range and the voltage pulse amplitude and width, a VCM ReRAM device can operate in two different modes, \ie, binary or analog \citep{Cueppers19_091105}.
In the analog mode, the applied pulses result in a gradual monotonous change of the device conductance, for both potentiation and depression. 
This operation mode is suitable for the implementation of STDP-type learning rules.
It is, however, characterized by a limited conductivity range, and the device switching characteristics may slowly drift away from the analog behavior to a more abrupt conductivity change.
In the binary mode, the conductivity can only be switched between two values, the LCS and the HCS. 
The switching between these two states occurs abruptly. 
In previous works, the abrupt, binary switching is achieved using single program pulses with a sufficiently large amplitude \citep{Cueppers19_091105}.
In contrast, here, we study the switching behavior of the device as a response to a certain number of pulses of smaller amplitudes.
As a response to these pulses, an internal state variable $N_\text{VO}$ gradually increases \citep{Fleck16_064015}.
Only when this $N_\text{VO}$ exceeds a certain threshold value, a thermal runaway condition is reached resulting in an abrupt switching event.
Due to intrinsic ReRAM device variabilities \citep{Fantini13_30}, the number of pulses to reach this thermal runaway conditions shows a strong device-to-device and cycle-to-cycle variation.
During the depression, the switching is intrinsically more gradual, due to the lack of an internal runaway mechanism as present for the potentiation operation.
Adding a series resistance (in or outside the device) can provide such runaway mechanism due to a voltage divider effect also in the RESET case \citep{Hardtdegen18_3229}. 
Hence, in both cases, the switching behavior can be summarized as follow: at first only a gradual change of the internal state variable $N_\text{VO}$ is observed, associated with only a minor change of the device conductivity, followed by a strong switching effect (large change of $N_\text{VO}$ as well as of the associated conductivity) when the internal state variable reaches a certain threshold.
This operation mode is of particular interest for this study, as it is similar to the structural STDP plasticity discussed and implemented in the original spiking TM model \citep{Bouhadjar22_e1010233}.
\par
Previous studies suggested both physics-based and phenomenological models for VCM type ReRAMs.
Physics-based models such as the JART model \citep{Bengel20_4618} capture detailed physical characteristics and predict their specific experimental behavior.
They require however long simulation time and lead to convergence issues. 
On the other hand, the more phenomenological
models give a high-level description of the operational characteristics, have good accuracy, are computationally less demanding, and can hence be combined with large-scale network models.
In this study, we opt for a phenomenological model to implement both the analog and the binary ReRAM device.
\par
The synapses are either potentiated or depressed by following learning rules similar to those outlined in the spiking TM model.
The learning rules are implemented by the control circuit (\cref{fig:controller_memristor}) as follows: the synapse is depressed slightly at every presynaptic spike and potentiated if a postsynaptic spike follows after a presynaptic spike.
In contrast to the original spiking TM model, synapses are potentiated by a fixed amount irrespective of the relative timing between the pre- and postsynaptic spikes.
The potentiation is however disabled if these spikes occur very close to each other within the interval [0, $\Delta{}T_\text{min}$]. 
This prohibits synchronously firing neurons from connecting to each other.
The control circuit further implements a homeostatic control mechanism (see \cref{sec:reram_sequence_learning}). 
\par
In the analog mode, the increment
\begin{equation}
\label{eq:analog_synapse}
\Delta G_{i,j}=
\begin{cases}
G_\text{max} \cdot \lambda_{+} \cdot \left(1 - \dfrac{G_{i,j}}{G_\text{max}}\right)^{\mu_{+}} + X & \text{for potentiation}\,\\[10pt] 
-G_\text{max}  \cdot \lambda_{-} \cdot \left(\dfrac{G_{i,j}}{G_\text{max}}\right)^{\mu_{-}} + X & \text{for depression}
\end{cases}    
\end{equation}
in the conductivity of the device (synapse) $j\to{}i$ following a potentiation or a depression event is modeled as in \citep{Fusi07_485}, but with an additional additive noise $X$.
For each synapse and for each potentiation and depression step, the noise $X \sim \mathcal{N}(0,\,\sigma_\text{w}^{2})$ is randomly and independently drawn from a normal distribution with zero mean and standard deviation $\sigma_\text{w}$.
The conductance $G_{i,j}$ evolves between a lower and an upper bound $G_\text{min}$ and $G_\text{max}$, and it is clipped at these boundaries, with learning rates $\lambda_{+}$ and  $\lambda_{-}$ and weight dependence exponents $\mu_{+}$ and $\mu_{-}$.
The conductance changes linearly with the internal state variable $N_\text{VO}$, thus no specification of the internal state variable is necessary.
The initial conductance $G_\text{min}=G_{ij}(0)$ is drawn for every new device from a uniform distribution in the interval [$G_{0,\text{min}},G_{0,\text{max}}$]. 
\par
For the binary switching behavior, we use a similar model as the structural STDP model proposed by \citet{Bouhadjar22_e1010233}.
The switching of the conductance between the LCS and the HCS is controlled by a permanence $P$.
The permanence plays the role of the internal state variable $N_\text{VO}$. If it is above a certain threshold $\theta_\text{P}$, the conductance $G_{i,j}$ is set to $G_\text{max}$, otherwise it is set to $G_\text{min}$:
\begin{equation}
  \label{eq:binary_synapse_thresholding}
  G_{i,j}(t) = \begin{cases}
  G_\text{max}  & \mbox{if}\ P_{ij}(t) \geq \theta_P  \\
  G_\text{min}  & \mbox{if}\ P_{ij}(t) < \theta_P. 
  \end{cases}
\end{equation}
Similar to the analog synapse, the initial conductance $G_\text{min}$ is drawn for every new device from a uniform distribution in the interval [$G_{0,\text{min}},G_{0,\text{max}}$].
At each potentiation or depression step, the permanence $P$ of the synapse $j\to{}i$ is incremented by an amount
\begin{equation}
\label{eq:binary_synapse}
\Delta P_{i,j}=
\begin{cases}
P_\text{max} \cdot \lambda_{+} \cdot \left(1 - \dfrac{P_{i,j}}{P_\text{max}}\right)^{\mu_{+}} + X & \text{for potentiation}\,\\[10pt] 
-P_\text{max} \cdot \lambda_{-} \cdot \left(\dfrac{P_{i,j}}{P_\text{max}}\right)^{\mu_{-}} + X & \text{for depression},
\end{cases}
\end{equation}
%
similar to the conductance increment of the analog synapse.
It has a lower and an upper bound $P_\text{min}$ and $P_\text{max}$ and it is clipped at these boundaries.
While the maximum permanences $P_\text{max}$ are identical for synapses, the minimal permanences $P_{\text{min},ij}$ are uniformly distributed in the interval $[P_{0,\text{min}},P_{0,\text{max}}]$.
\par
In addition to the write noise introduced by means of the variable $X$, both the analog and the binary synapse models incorporate a read noise.
At each presynaptic spike of neuron $j$, a noisy component Z is added to the synaptic current
\begin{equation}
  I_{i,j}(t)=(G_{i,j}(t) + Z)\cdot V_\text{read}(t)=\overline{G}_{i,j}(t)\cdot V_\text{read}(t),
\end{equation}
of neuron $i$, where $Z\sim\mathcal{N}(0,\,\sigma_\text{r}^{2})$ is randomly and independently drawn from a normal distribution with zero mean and standard deviation $\sigma_\text{r}$, and $V_\text{read}(t)$ is the applied voltage. In the course of this article, we use $\overline{G}$ to denote the conductance incorporating both the read and the write noise.
\par
\begin{figure}[!h]
  \centering
  \includegraphics{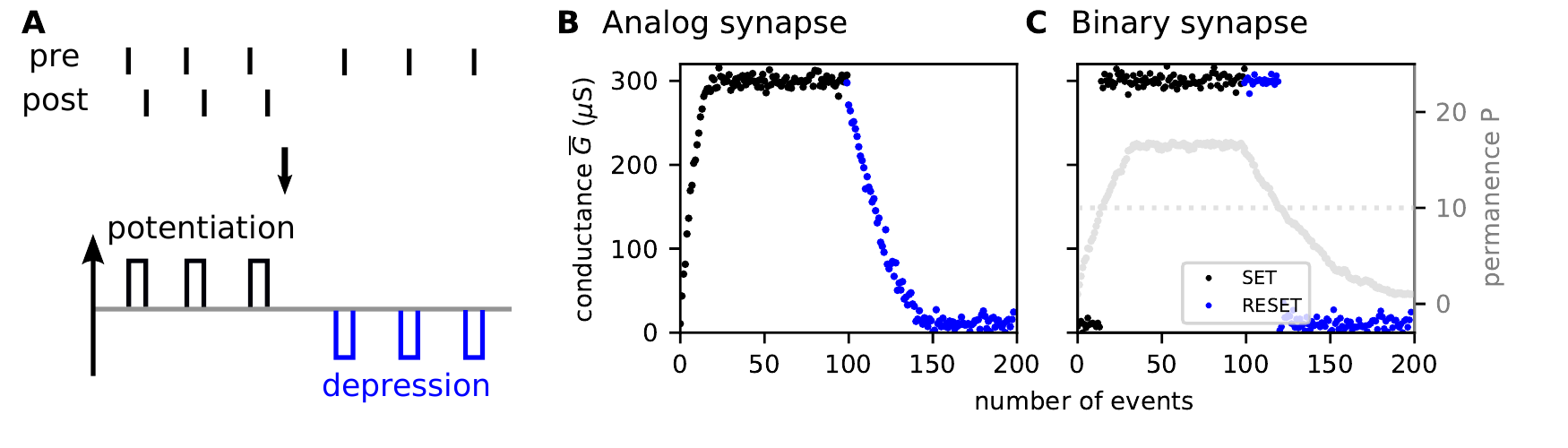}
  \caption{%
    \textbf{Intrinsic dynamics of the ReRAM model (simulation results).}
    \panel{A} Sketch of the experimental protocol and mapping of pre- and postsynaptic spike timing (top) to
    the corresponding SET (potentiation; black) and RESET (depression; blue) operations (bottom).
    Evolution of the conductance $\overline{G}$ in response to $100$ SET (potentiation; black) updates, followed by $100$ RESET (depression; blue) updates, for the analog (\panellabel{B}) and the binary ReRAM model (\panellabel{C}). 
    In C, the permanence of the binary device is plotted in grey.
    Parameters:
    learning rates 
    $\lambda_{+}=\lplusa$, $\lambda_{-}=\lambda_{+}/3$ (analog synapse), 
    $\lambda_{+}=\lplusb$, $\lambda_{-}=\lambda_{+}/3$ (binary synapse),
    weight dependence exponents $\mu_{+}=\mplus$, $\mu_{-}=\mminus$, and noise amplitudes $\sigma_{r}=\sr$, $\sigma_{w}=\sw$.
    For remaining parameters, see \cref{tab:Model-parameters}.
  }
\label{fig:intrinsic_memristive_dynamics} 
\end{figure}
\begin{figure}[!h]
  \centering
 \includegraphics{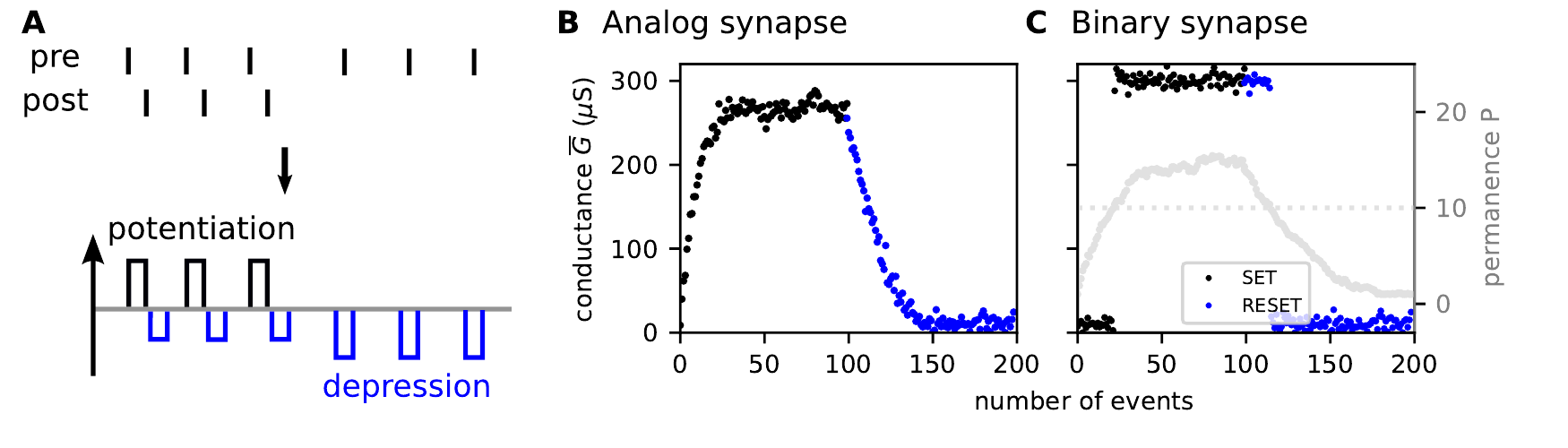}
  \caption{%
    \textbf{Dynamics of the ReRAM model in the spiking TM context.}
    Same figure arrangement as in \cref{fig:intrinsic_memristive_dynamics}. In the context of the spiking TM model, each potentiation pulse is accompanied by a smaller depression pulse.
  }
  \label{fig:memristive_dynamics}
\end{figure}

\Cref{fig:intrinsic_memristive_dynamics} shows an exemplary switching behavior of the analog and binary synapse models for a specific set of parameters using $100$ consecutive potentiation (i.e., SET) and depression (i.e., RESET) updates. We choose different learning rates ($\lambda_{+}$ and $\lambda_{-}$) for the two types of devices such that they switch from the LCS to the HCS (and back) at about the same number of updates.
Under normal operation of the spiking TM model, a potentiation update is always followed by a small depression (\cref{fig:memristive_dynamics}A). 
In the case of the analog synapse, the total synaptic growth in the absence of noise is therefore governed by
\begin{equation}
  \label{eq:learning_effect}
  \Delta G_{i,j} =  G_\text{max} \left[ \lambda_{+} \cdot \left(1 - \dfrac{G_{i,j}}{G_\text{max}}\right)^{\mu_{+}}  - \lambda_{-} \cdot \left(\dfrac{G_{i,j}}{G_\text{max}}\right)^{\mu_{-}}\right].
\end{equation}
The stationary solution of the device conductance (fixed point) $G^*$, obtained by setting $\Delta G_{i,j}=0$, is always below the maximum conductance $G_\text{max}$ (\cref{fig:memristive_dynamics}B).
The permanence of the binary synapse is subject to this effect, too. After a number of potentiation steps, it reaches a value $P^*$ smaller than $P_\text{max}$ (see \cref{fig:memristive_dynamics}C). According to \cref{eq:binary_synapse_thresholding}, the conductance can however still assume $G_\text{max}$. Only if the depression is too strong, the device may not reach the maturity threshold $\theta_P$, and thus not switch to the HCS.
\par
In the next sections, we evaluate the effects of different characteristics of the analog and the binary switching dynamics such as the weight dependence of the device update ($\mu_{+}$, $\mu_-$), the conductance range ($G_\text{min}$, $G_\text{max}$), the learning rates ($\lambda_{+}$, $\lambda_{-}$), as well as the write and the read variability ($\sigma_\text{w}$, $\sigma_\text{r}$) on the learning process of the spiking TM model.


\subsection{A spiking neural networks with ReRAM synapses successful at sequence prediction}
\label{sec:reram_sequence_learning}

Sequence learning and prediction are principal computations performed by the brain and have a number of potential technological applications.
The study by \cite{Bouhadjar22_e1010233} proposed a model of this type of computation known as the spiking temporal memory (spiking TM) model. 
It consists of a sparsely and recurrently connected network of spiking neurons and learns sequences continuously in an unsupervised manner by means of known biological plasticity mechanisms.
After learning, the network successfully predicts and recalls complex sequences in a context-specific manner, and signals anomalies in the data.
\par
\begin{figure}
    \centering
    \includegraphics{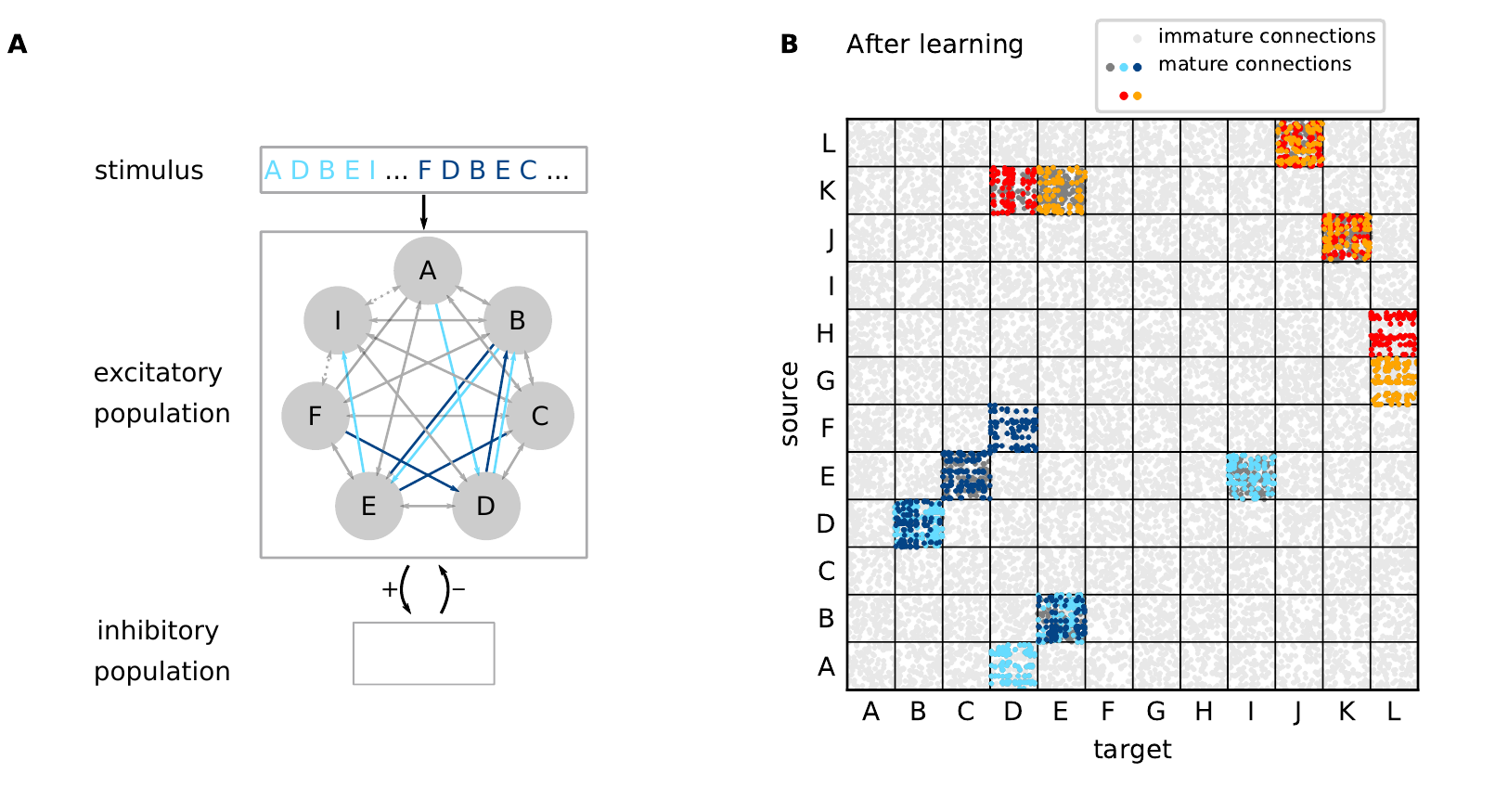}
    \caption{%
    \textbf{Network structure.} 
      \panel{A} Sketch of the model architecture composed of a randomly and sparsely connected recurrent network of excitatory and inhibitory neurons.    
      The excitatory neuron population is subdivided into subpopulations according to stimulus preference (gray circles).
      During learning, sequence specific, sparsely connected subnetworks with mature synapses are formed (light and dark blue arrows).
      For the example discussed in the main text and in panel B, the network learns four high-order sequences \seq{A,D,B,E,I}, \seq{F,D,B,E,C}, \seq{H,L,J,K,D} and \seq{G,L,J,K,E}. In panel A, only two of them are depicted for clarity. The gray dashed lines depict the existence of further subpopulations, which are not shown in the sketch.
      \panel{B} Connectivity matrix of excitatory neurons after learning. 
      Target and source neurons are grouped into stimulus-specific subpopulations (``A'',\ldots,``F''). 
      During the learning process, subsets of connections between subpopulations corresponding to subsequent sequence elements become mature and effective (\seq{A,D,B,E,I}: light blue, \seq{F,D,B,E,C}: dark blue, \seq{H,L,J,K,D}: red, \seq{G,L,J,K,E}: orange). 
      Immature synapses are marked by light gray dots.
      Dark gray dots in panel B correspond to mature connections between neurons that remain silent after learning.
      Only $1\%$ of immature connections are shown for clarity.
    }
    \label{fig:network_structure_activity}
\end{figure}

We briefly describe here the main mechanisms and principles of the spiking TM model. 
For an in-depth analysis, we refer readers to \citep{Bouhadjar22_e1010233}.
The model is composed of a $\NE$ excitatory (``E'') and $\NI$ inhibitory (``I'') neurons, which are randomly and sparsely connected. 
Excitatory neurons are subdivided into $M$ subpopulations of neurons, each representing a sequence element and sharing the same stimulus preference (\cref{fig:network_structure_activity}A).
Excitatory neurons are recurrently connected to the inhibitory neurons implementing a winner-take-all (WTA) mechanism. 
We model neurons using leaky integrate-and-fire dynamics. 
Excitatory neurons are additionally equipped with nonlinear dendrites mimicking dendritic action potentials (dAPs).
We model the dAPs as follows: if the dendritic current exceeds a threshold $\theta_{\text{dAP}}$, it is instantly set and clamped to the dAP plateau current $I_\text{dAP}$ for a period of duration $\tau_\text{dAP}$. 
The dAP threshold is chosen such that the co-activation of $\gamma$ presynaptic neurons reliably triggers a dAP in the target neuron:
\begin{equation}
  \label{eq:fix_dAP}
  \theta_{\text{dAP}} = G_{+} \cdot \gamma \cdot p.     
\end{equation}
In the case of the analog synapse, $G_{+}$ is taken to be the steady-state conductance $G^*$, and in the case of the binary synapse, it is taken to be $G_\text{max}$.
In a previous study, we show that it is also possible to model the dendrites using leaky integrate-and-fire dynamics \citep{Bouhadjar19_ICONS}.
In addition to the dendritic input, the excitatory neurons are equipped with additional inputs from external and inhibitory sources.
Inhibitory neurons have only a single excitatory input.
The synapses between excitatory neurons are plastic evolving according to the analog or the binary ReRAM models described in \cref{sec:reram_model}.
A homeostatic component further controls the synaptic growth: if the dAP activity, i.e, the number of generated dAPs in a certain time window, is below a target $z^*$, the synaptic weight is increased, otherwise, it is decreased (see \cref{sec:methods}).
\par
%
During the learning process, the network is repeatedly presented with a given ensemble of sequences.
Before learning, presenting a sequence element causes all neurons in the respective subpopulation to fire, except the subpopulation representing the first sequence element, where only a random subset of neurons is activated.
The repeated presentation of the sequences strengthens the connections between the subpopulations representing subsequently presented elements. 
After sufficient learning, the activation of a subpopulation by an external input causes a specific subset of neurons in the following subpopulation to generate dendritic action potentials (dAPs) resulting in a long-lasting depolarization of the somata.
Neurons that generate dAPs signal the anticipated sequence element and are thus referred to as predictive neurons.
When receiving an external input, predictive neurons fire earlier as compared to non-predictive neurons.
If a certain subpopulation contains a sufficient number of predictive neurons, their advanced spike initiates a fast and strong inhibitory feedback to the entire subpopulation, ultimately suppressing the firing of the non-predictive neurons.
The randomness in the connectivity supplemented by the homeostatic control enables the generation of sequence-specific sparse connectivity patterns between subsequently activated neuronal subpopulations (\cref{fig:network_structure_activity}A,B).
For each pair of sequence elements in a given sequence ensemble, there is a unique set of postsynaptic neurons generating dAPs.
Consequently, after learning in response to the presentation of a sequence element, the network predicts in a context-dependent manner the next element in the sequence by activating the dAPs of the corresponding subpopulation (\cref{fig:network_structure_activity}C,D).
\par
\begin{figure}[!h]
  \centering
  \includegraphics{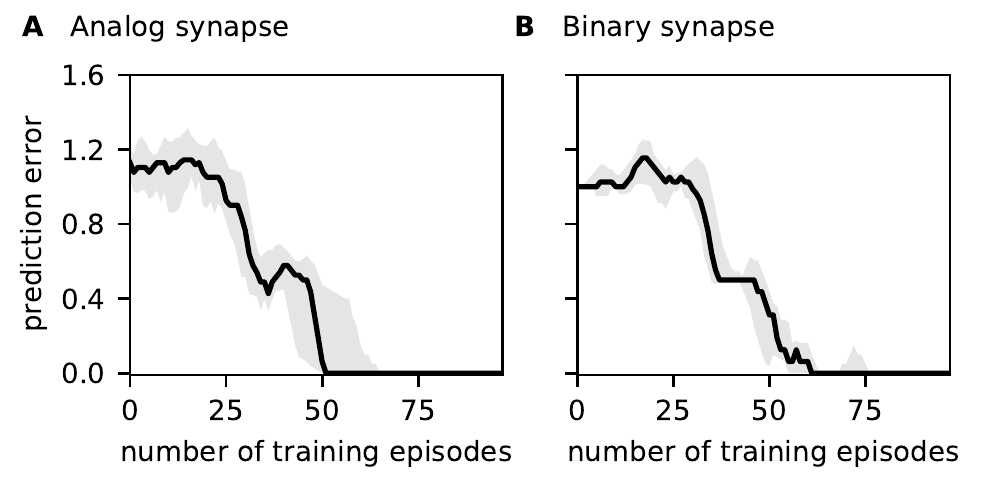}
  \caption{%
    \textbf{Prediction error.}
    Dependence of the prediction error on the number of training episodes for the network with analog synapses (\panellabel{A}) or binary synapses (\panellabel{B}).
    Curves and error bands indicate the median as well as the $5\%$ and $95\%$ percentiles across an ensemble of $5$ different network realizations, respectively.
    Same parameters as in \cref{fig:intrinsic_memristive_dynamics}.
  }
\label{fig:prediction_error} 
\end{figure}

Here, we study the prediction performance for the network with either the binary or the analog ReRAM synapses (\cref{fig:prediction_error}).
We use the synaptic parameters fitted from the exemplary data discussed in \cref{sec:reram_model}.
To quantify the sequence prediction performance, we repetitively stimulate the network using the same set of sequence \seq{A,D,B,E,I}, \seq{F,D,B,E,C}, \seq{H,L,J,K,D}, \seq{G,L,J,K,E} and assess the prediction error by comparing the the anticipated next sequence element with the correct one \citep{Bouhadjar22_e1010233}.
To ensure the performance results are not specific to a single network, the evaluation is
repeated for a number of randomly instantiated network realizations with different initial connectivities. 
After each new network instantiation, the initial prediction error is at $1$ (\cref{fig:prediction_error}).
With an increasing number of training episodes, the prediction error for both networks with either the binary or the analog synapses decreases to zero as both networks learn the sequences and develop context-dependent pathways between successive sequence elements.

\subsection{Influence of device characteristics on prediction performance}

ReRAM devices reported in the literature exhibit different nonidealities, including 1) limited precision or the number of synaptic levels; 2) limited dynamic range; 3) dependence of the synaptic updates on the weight, also known as synaptic nonlinearity; 4) device variability, including read and write variability \citep[see][for an overview]{Zhao20_011301}.
In this section, we study how these nonidealities affect the prediction performance in the spiking TM model.

\subsubsection{Optimal prediction performance obtained for a broad range of on-off ratios and learning rates}

The dynamic range is defined as the on-off ratio between the minimum ($G_\text{min}$) and the maximum conductance ($G_\text{max}$).
Most ReRAM devices exhibit an on-off ratio in a range of $2$x to $>$ $10^4$x \citep{Hong18_8720}.
Within the minimum and the maximum conductance, the
synaptic precision or the number of synaptic steps is limited.
In the synapse model proposed in \cref{sec:reram_model}, we can influence the number of steps by changing parameters such as the learning rates ($\lambda_{+}$, $\lambda_{-}$), weight dependence exponents ($\mu_{+}$,$\mu_{-}$), or the on-off ratio.
Given the difficulty to derive an analytical solution of the number of steps as a function of these parameters, we restrict the scope of the study in this section to investigating the influence of different learning rates and on-off ratios on the prediction performance.
%
\par
\begin{figure}[!h]
  \centering
  \includegraphics{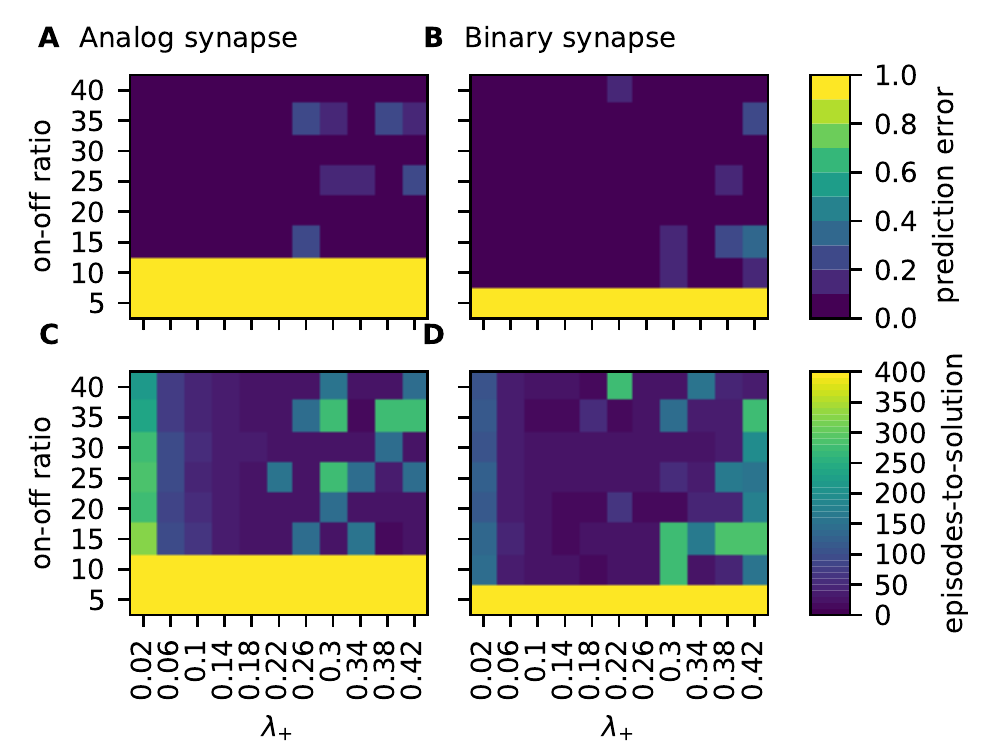}
  \caption{%
    \textbf{Effect of the on-off ratio and the learning rate on the prediction performance.}
    Dependence of the prediction error and episodes-to-solution on the on-off ratio and the learning rate shown for the network with either analog (\panellabel{A,C}) or binary synapses (\panellabel{B,D}).
    Data depicts the median across an ensemble of $5$ different network realizations.
    Parameters: depression learning rate $\lambda_{-}=\lambda_{+}/3$, weight dependence exponents $\mu_{+}=\mu_{-}=\mplus$, and variability amplitudes $\sigma_{r}=\sr$, $\sigma_{w}=\sw$.
    For remaining parameters see \cref{tab:Model-parameters}.
  }
\label{fig:mat_prediction_performance} 
\end{figure}

We vary the on-off ratio between $5$ and $40$ by keeping $G_\text{min}$ fixed and varying $G_\text{max}$.
As $G_\text{min}$ is drawn from a uniform distribution in the interval [$G_{0,\text{min}}$, $G_{0,\text{max}}$], we compute the on-off ratio as $G_\text{max}/G^{*}_\text{min}$, where $G^{*}_\text{min}=(G_{0,\text{max}}+G_{0,\text{min}})/2$.
As we change $G_\text{max}$, we modify the dAP threshold, see \cref{eq:fix_dAP}.
In addition, we vary the learning rate between $2\%$ and $42\%$ (\cref{fig:mat_prediction_performance}).
Parameters such as the read and write variability and the weight dependence exponents are taken from the exemplary data presented in \cref{sec:reram_model}.
We study the influence of the variability and the dependence of the synaptic updates on the weight more systematically in the upcoming sections.
For the analog synapse, the prediction error converges to zero for an on-off ratio between $15$ and $40$ and for a learning rate between $2\%$ and $18\%$.
For the binary synapse, successful learning is obtained for an on-off ratio between $10$ and $40$ and for a learning rate between $2\%$ and $18\%$ (\cref{fig:mat_prediction_performance}A,B).
For learning rates above $18\%$, the prediction performance becomes less stable with sudden failures for some network realizations. 
While decreasing the learning rate yields minimum prediction error, the number of episodes-to-solution tends to increase as either the conductances or permanences need more learning steps to reach their maximum value (\cref{fig:mat_prediction_performance}C,D).
The learning in the network with binary synapses is faster due to the internal dynamics of binary synapses, which has faster switching dynamics compared to analog synapses: the permanence takes less number of update steps to reach the maturity threshold ($\theta_\text{P}$) compared to the number of update steps the conductance of the analog synapses require to go from the LCS to the HCS.
\par
In general, the on-off ratio in the spiking TM network is limited due to the following:
the transition of the network activity from being initially non-sparse to becoming sparse after learning requires initial small conductances to avoid spurious activation of the dAPs, but high conductances after learning to allow the sparse set of active neurons to generate the dAP reliably.
If the on-off ratio is too small this distinction between high and small conductances cannot be realized.
Moreover, for successful learning, the network with analog synapses requires a higher on-off ratio compared to the network with binary synapses.
This is due to the effect described in \cref{sec:reram_model} below equation \cref{eq:learning_effect}, which prohibits the conductance from reaching $G_\text{max}$. Therefore, the effective on-off ratio is reduced.
The learning mechanisms of the spiking TM also limit the range of possible learning rates.
Increasing the learning rate bears the risk that a large fraction of neurons reaches the dAP threshold at the same time. 
The WTA mechanism selects then all neurons that generate dAP to become active. 
This leads to a loss of sparseness, which results in impairing the prediction performance.
Decreasing the learning rate considerably is also not ideal as the network would learn very slowly.

\subsubsection{Resilience of the model against weight dependent updates}

\begin{figure}[!h]
  \centering
  \includegraphics{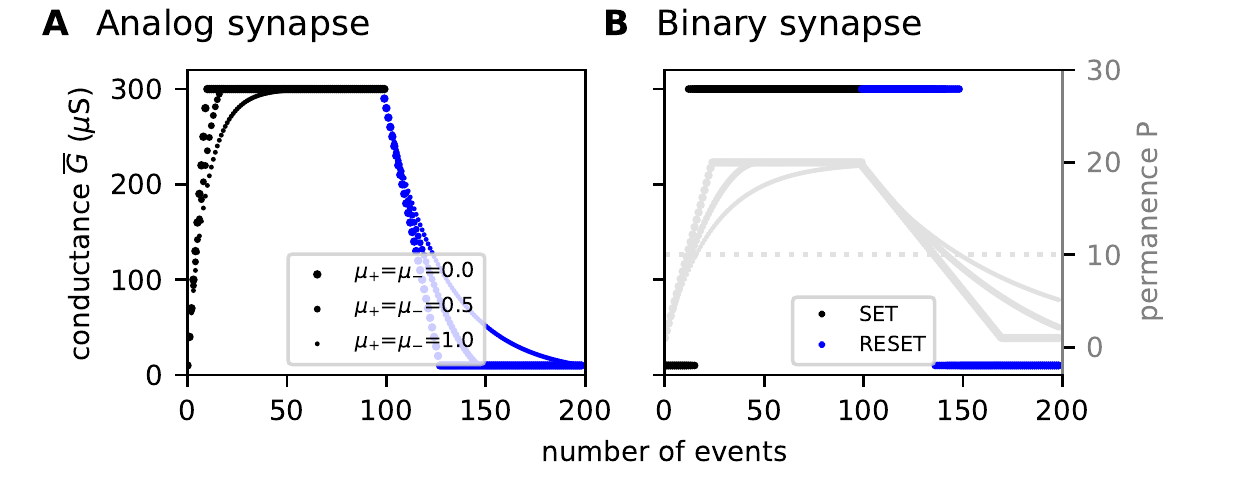}
  \caption{%
    \textbf{An exemplary potentiation and depression curves depicting different levels of weight dependence updates.}
    Dependence of the conductance $\overline{G}$ on the number of either SET (potentiation; black) or RESET (depression; blue) events as well as on different weight dependence exponents $\mu_{+}=\mu_{-}=0$ (large dot), $\mu_{+}=\mu_{-}=0.5$ (small dot), and $\mu_{+}=\mu_{-}=1$ (tiny dot) plotted for the analog (\panellabel{A}) and the binary ReRAM models (\panellabel{B}).
    In B, the permanence of the binary device is plotted in grey.
    Parameters: 
    learning rates 
    $\lambda_{+}=\lplusa$, $\lambda_{-}=\lambda_{+}/3$ (analog synapse), 
    $\lambda_{+}=\lplusb$, $\lambda_{-}=\lambda_{+}/3$ (binary synapse),
    and variability amplitudes $\sigma_{w}=0$, $\sigma_{r}=0$.
    For remaining parameters see \cref{tab:Model-parameters}.
  }
\label{fig:nonlinear_reram_dynamics} 
\end{figure}

The evolution of the conductance of realistic analog synapses grows in a nonlinear manner as a function of the potentiation and depression updates.
The synapse model in \cref{sec:reram_model} captures this effect via the weight dependence factor controlled by the exponents ($\mu_{+}$, $\mu_{-}$).
During the potentiation process, the conductance tends to change rapidly at the beginning but saturates at the end of the process (see \cref{fig:nonlinear_reram_dynamics}A).
Similar behavior is also observed during the RESET.
The potentiation and depression operations have, however, different dependencies on the device conductance.
For high conductances, the potentiation increments are much smaller than the depression decrements.
This asymmetry in the behavior can be further enhanced if the learning rates are different during the potentiation and depression operations.
Similarly, it is reasonable to assume that for the binary synapses the evolution of the permanence may exhibit a nonlinear dependence on the synaptic updates and an asymmetric behavior between the potentiation and depression dynamics (\cref{fig:nonlinear_reram_dynamics}B). 
\par
%
Here, we first evaluate how the asymmetry in the learning rates between the potentiation and depression operations ($\lambda_{+}$ and $\lambda_{-}$) affects the prediction performance.
To study this effect, we fix $\lambda_{+}$ and vary $\lambda_{-}$ with the state dependence exponents $\mu_{+}$ and $\mu_{-}$ being set to zero.
The prediction error remains high if $\lambda_{-}\geq\lambda_{+}$ (see \cref{fig:supp_asy_prediction_error}).
This is due to the plasticity dynamics of the spiking TM model: the potentiation operation is applied only when the postsynaptic spike follows after the presynaptic spike, in contrast, the RESET operation is applied every time the presynaptic neuron generates a spike.
Therefore, for effective synaptic growth, the potentiation needs to be stronger than depression.
%
\par
\begin{figure}[!h]
  \centering
  \includegraphics{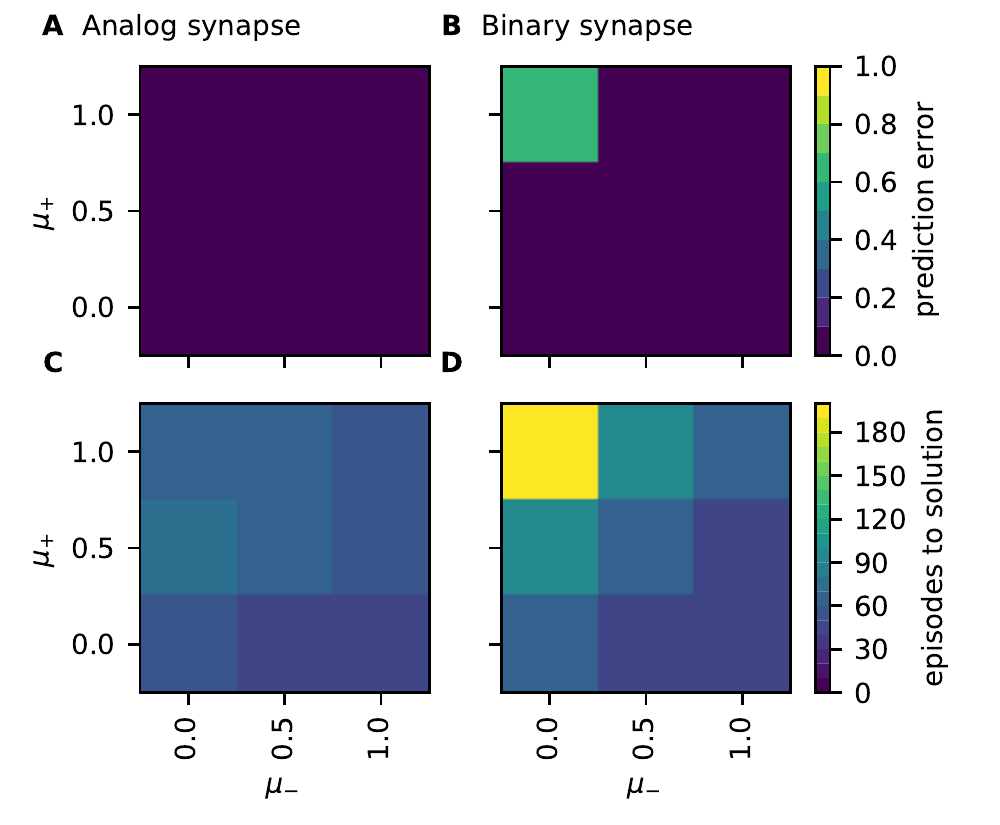}
  \caption{%
    \textbf{Effect of the weight dependence update on the prediction performance.}
    Dependence of the prediction error and episodes-to-solution on the weight dependence exponents for both potentiation and depression ($\mu_{+}$ and $\mu_{-}$) shown for the networks with either  analog (\panellabel{A,C}) or binary synapses (\panellabel{B,D}).
    Data depicts the median across an ensemble of $5$ different network realizations.
    Parameters: 
    learning rates 
    $\lambda_{+}=\lplusa$, $\lambda_{-}=\lminusa$ (analog synapse), 
    $\lambda_{+}=\lplusb$, $\lambda_{-}=\lminusb$ (binary synapse),
     and variability amplitudes 
    $\sigma_{w}=\sw$, $\sigma_{r}=\sr$.
    For remaining parameters see \cref{tab:Model-parameters}.
  }
\label{fig:nonlinear_prediction_error} 
\end{figure}

We assess, next, the prediction performance as a function of different weight dependence exponents for both potentiation and depression ($\mu_{+}$ and $\mu_{-}$, respectively, see \cref{fig:nonlinear_prediction_error}).
The results show that this latter has mild effects on the prediction error (see \cref{fig:nonlinear_prediction_error}A,B).
For larger values of $\mu_{+}$, the learning speed slows down as it takes longer for either the conductance or the permanence to reach their maximum values (see \cref{fig:nonlinear_prediction_error}C,D).
Decreasing $\mu_{-}$ makes learning faster again as the depression becomes weaker compared to the potentiation.
In the binary case, the steady-state permanence $P^*$ may end up below the maturity threshold $\theta_\text{P}$ such that the synapses can mature only due to the noise.
The learning is therefore slowed down for large values of $\mu_{+}$ or even unsuccessful if the devices do not switch to the HCS. 
In the model, $\theta_\text{P}$ could be adjusted to $P^*$ (similarly to adjusting $\theta_\text{dAP}$ to $G^*$ in the analog synapse; see above).
In this case, the learning in the analog and the binary networks may be similarly fast. 
In the physical device, however, the maturity threshold $\theta_\text{P}$ can hardly be changed.

\subsubsection{Resilience of the model against variability}

\begin{figure}[!h]
  \centering
  \includegraphics{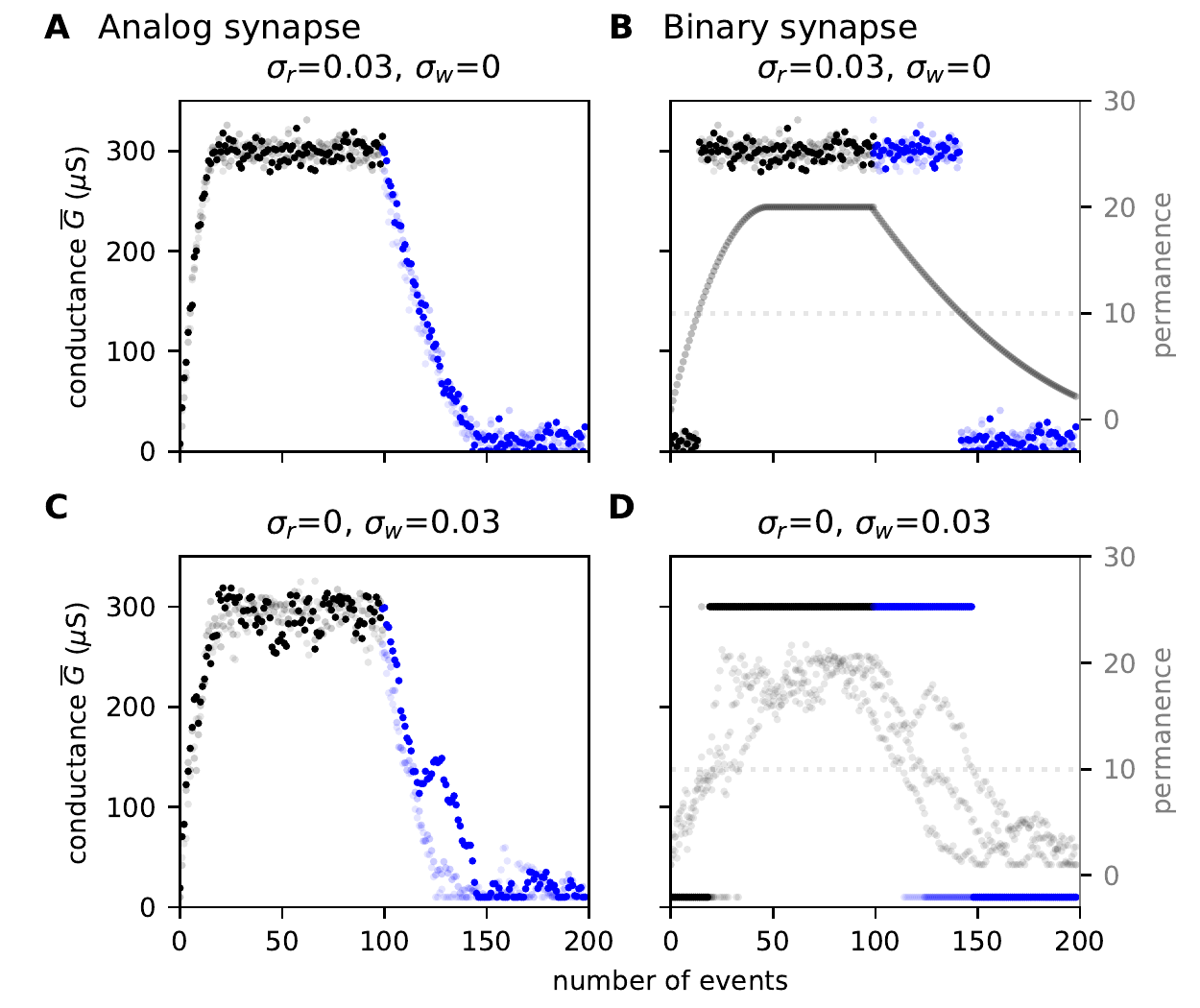}
  \caption{%
    \textbf{An exemplary potentiation and depression curves in the presence of read and write variability.}
    Dependence of the conductance $\overline{G}$ on the number of either potentiation (black) or depression (blue) events in the presence of read noise ($\sigma_\text{r}=\sr$, \panellabel{A,B}) or write noise ($\sigma_\text{w}=\sw$, \panellabel{C,D}) plotted for the analog (B) and the binary ReRAM models (C).
    In B and D, the permanence of the binary device is plotted in gray.
    Parameters: learning rates 
    $\lambda_{+}=\lplusa$, $\lambda_{-}=\lambda_{+}/3$ (analog synapse), 
    $\lambda_{+}=\lplusb$, $\lambda_{-}=\lambda_{+}/3$ (binary synapse),
    and weight dependence exponents $\mu_{+}=\mu_{-}=\mplus$.
    For remaining parameters see \cref{tab:Model-parameters}.
  }
\label{fig:var_reram_dynamics}
\end{figure}

The resistive switching process of ReRAM devices involves the drift and diffusion of the oxygen vacancies.
This phenomena is highly stochastic and shows considerable variation from device to device, and even from pulse to pulse within one device \citep{Zhao20_011301}.
Further, even when no switching occurs, the oxygen vacancies exhibit random microscopic displacements resulting in the read variability.
%
In our work, we capture these effects by the read and write variability introduced in \cref{sec:reram_model}.
The influence of the read and write variability on the conductance curves are illustrated for both the analog and binary synapses in \cref{fig:var_reram_dynamics}.
For different trials, the write variability results in different conductance trajectories as a function of the applied potentiation or depression events.
The read variability, on the other hand, causes only a jitter in the conductance curves.
%
\par
\begin{figure}[!h]
  \centering
  \includegraphics{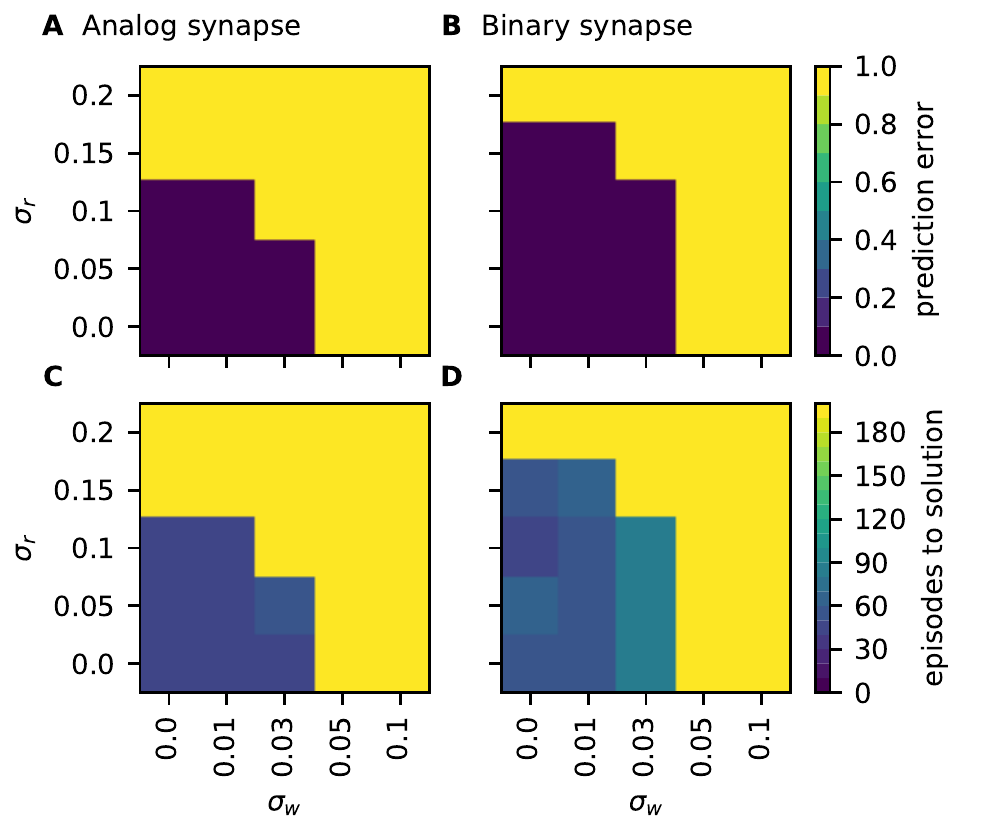}
  \caption{%
    \textbf{Effect of the variability on the prediction performance.}
    Dependence of the prediction error and episodes-to-solution on the read and write variability $\sigma_\text{r}$, $\sigma_\text{w}$, shown for the networks with either analog (\panellabel{A,C}) or  binary synapse (\panellabel{B,D}).
    Data depicts the median across an ensemble of $5$ different network realizations.
    Parameters: learning rates 
    $\lambda_{+}=\lplusa$, $\lambda_{-}=\lambda_{+}/3$ (analog synapse), 
    $\lambda_{+}=\lplusb$, $\lambda_{-}=\lambda_{+}/3$ (binary synapse),
    and weight dependence exponents $\mu_{+}=\mu_{-}=\mplus$.
    To gain more robustness with respect to the variability (decrease false negatives), we decrease $\theta_\text{dAP}$ by $10\%$ as compared to the default value described in \cref{eq:fix_dAP}.
    For remaining parameters see \cref{tab:Model-parameters}.
  }
\label{fig:var_prediction_error} 
\end{figure}

To study how the variability influences the prediction performance, we assess the prediction error and episodes-to-solution for different magnitudes of the read and write variability $\sigma_\text{r}$ and $\sigma_\text{w}$, respectively.
Both networks with either analog or binary synapses allow similar read and write noise levels, with the binary synapse being slightly more resilient toward the read noise  (\cref{fig:var_prediction_error}A,B).
In both cases, the write noise is more detrimental as it accumulates across the different learning episodes and can therefore have a higher impact on the learning performance. 
The read noise tends to average out as it is independent across the learning episodes.
Overall, increasing the read or write variability beyond what is acceptable leads to a spurious activation of the dAPs, i.e., predictions, and a decline in the prediction performance.
Concerning the learning speed, the number of episodes-to-solution is similar for the different variability levels where the learning is successful (\cref{fig:var_prediction_error}C,D).

\subsubsection{Robustness of the model against synaptic failure}

\begin{figure}[!h]
  \centering
  \includegraphics{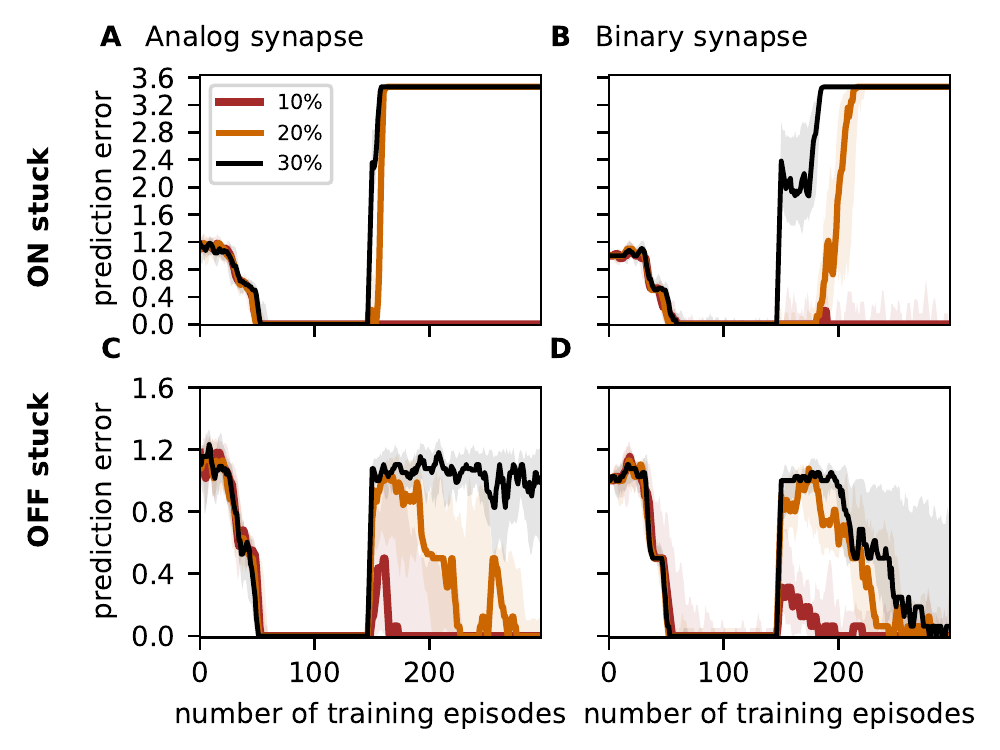}
  \caption{%
    \textbf{Effect of synaptic failure on the prediction performance.}
    Dependence of the prediction error on the number of training episodes and different levels of synaptic failure (red \sfl\%, orange \sfi\%, black \sfh\%) shown for both the analog (\panellabel{A}, \panellabel{C}) and the binary synapse (\panellabel{B}, \panellabel{D}).
    We implement the synaptic failure by fixing a random selection of synapses to be stuck at the HCS (ON stuck; A and B) or stuck at the LCS (OFF stuck; C and D).
    The synaptic failure is introduced at episode $150$.
    Curves and error bands indicate the median as well as the $5\%$ and $95\%$ percentiles across an ensemble of $5$ different network realizations, respectively.
    Same parameters as in \cref{fig:intrinsic_memristive_dynamics}.
  }
\label{fig:synaptic_failure} 
\end{figure}

When operating ReRAM devices, they risk failing by getting trapped in the HCS even after applying voltage pulses with the appropriate magnitude across them \citep{Kumar17_10}. 
To study how synaptic failure affects the prediction performance, we first train the network till it reaches zero prediction error (after $150$ episodes in \cref{fig:synaptic_failure}).
Then, the conductance of a random fraction of synapses is set to the HCS. 
We quantify the level of synaptic failure by the ratio between the number of failed synapses and the total number of existing synapses.
In the spiking TM model, a neuron may falsely generate a dAP if a sufficient number of its synapses are randomly switched to the HCS (this number can be approximated as the ratio $\theta_\text{dAP}/G_\text{max}$, where $\theta_\text{dAP}$ is the dAP threshold and $G_\text{max}$ is the maximum conductance). This may result in generating false positives and thus an increase in the prediction error.
This is confirmed by our results presented in \cref{fig:synaptic_failure}A,B.
At up to $\sfl\%$ synaptic failure no impact is observed on the prediction performance (\cref{fig:synaptic_failure}A,B).
At greater than $\sfl\%$ synaptic failure the performance of the network declines and does not recover.
\par
In a second experiment, instead of turning a selection of random synapses to the HCS, we turn them to the LCS.
For the different levels of synaptic failures, the performance of the network initially declines.
Due to the failing synapses, which are stuck at the LCS, the neurons in certain subpopulations do not receive enough current and are thus not able to generate dAPs, i.e., make predictions.
After further training episodes, the prediction errors converge back to zero as the network relearns using other synapses (\cref{fig:synaptic_failure}C,D). 
At greater than $\sfi\%$ synaptic failure the performance does not recover due to the absence of alternative connections to form sequence-specific pathways.

\section{Discussion}

\subsection{Summary}

In this work, we demonstrate that the learning rules of the spiking temporal memory (spiking TM) model proposed by \cite{Bouhadjar22_e1010233} can be realized using memristive dynamics.
We investigate this for a particular type of memristive device known as VCM ReRAM \citep{Waser12_book}.
To this end, we show that the spiking TM retains high prediction performance for a broad range of on-off ratios and learning rates. 
The model is resilient toward the write and read variability as well as the dependence of the synaptic updates on the weight. 
Moreover, our results show that the VCM-type ReRAM device can be operated either in the binary or the gradual switching regime without performance loss.
This is in line with the original spiking TM implementation, which shows that the learning rule can either be implemented using structural plasticity where the weight abruptly changes between two levels or a conventional form of STDP where the weight gradually changes until it saturates.
%
This suggests that the intrinsic dynamics of the VCM ReRAM capture not only synaptic properties of biological synapses such as the variability and the dependence of the synaptic updates on the weight but also can implement known forms of plasticity in the neuroscientific literature.
%
Our study, therefore, ties further principles from the biological and the memristive synapses.  
%

\subsection{Relationship to previous models}

In artificial neural networks trained by gradient-based approaches, ReRAM nonidealities can severely undermine the overall performance \citep{Fouda20_499}. 
Due to the ReRAM variability, devices can be hardly programmed to a desired state, and the asymmetry in the conductance change can affect the propagation of the gradient and lead to performance loss.
Correcting for these nonidealities can be costly and may require additional circuitry \citep{Chen15_IEEE, Agarwal16, Hong18_8720, Yu18_260, Ambrogio18_60, Adnan21_1}.
%
%
We know that biological neuronal networks carry out accurate computations despite their synaptic nonideal characteristics such as variability.
This suggests the existence of biological principles accommodating that, which we need to understand and port to successfully implement neuromorphic hardware.
The spiking TM and other brain-inspired self-organizing networks \citep{Lazar09, Yi22_rain} suggest a set of biological concepts that might be at the heart of brain processing capabilities.
For instance, the highly sparse connectivity and activity of the spiking TM are observed in biological networks, and they are essential for increasing the capacity of the system and decreasing energy consumption. 
%
\par
There are a number of biologically motivated sequence learning models that are closely related to the spiking TM, such as the self-organizing recurrent neural network model \citep[SORN,][]{Lazar09}.
Recent work incorporated memristive dynamics into the synapses and neurons of the SORN model and showed that it retains successful performance \citep{Payvand22_5793}.
The authors studied the role of variability and showed that it can improve prediction performance.
However, the other memristive nonidealities were not studied systematically. It remains also to be investigated whether the model can learn high-order sequences similar to the ones presented in our work.

\subsection{Outlook}
Neuromorphic hardware that relies on components implemented in the analog domain is noisy and heterogeneous, similar to real brains \citep{Zhu20_011312}. 
To date, there are only speculations on how the brain contributes to sensible and reliable behavior in the face of these imperfections.
By using neuromorphic hardware as a test substrate, we expect to gain new insights into the neural principles that solve this issue.
In this study, it was apparent that ReRAM devices share the same characteristics as biological synapses including the weight dependence of the synaptic updates, limited dynamics range, and variability.
Throughout the work on these neuromorphic systems, we can develop intuitions of how the biological synapse exploits these different characteristics.
For instance, similar to biological synapses, ReRAM devices are characterized by a read and write variability.
In the biological system, the read variability is analogous to the randomness in the vesicle release, and the write variability corresponds to the randomness in the density regulation of the postsynaptic receptors.
So far, it is not clear how these different characteristics contribute to the learning dynamics in the biological system.
Neuromorphic hardware can provide an environment where this question can be studied.
\par
In this work, we show that the model is resilient toward synaptic variability.
Other works show that synaptic variability can even have a computational benefit \citep{Dalgaty21_151}.
For example, in probabilistic computing frameworks, the variability is considered a prerequisite for efficient probabilistic inference \citep{Buesing11, Suri13_2402, Maass14_860, Neftci16_241, Dutta22_2571}.
It allows the system to explore the state space and come up with an estimate of how likely is each solution.
Similarly, a recent extension of the spiking TM model shows that the model can learn to replay probabilistic sequences using noise \citep{Bouhadjar22_arxiv}.
The study however explored only correlated additive noise.
Future work can shed light on the possibility of using ReRAM noise, \ie, multiplicative noise, as an alternative.
\par

%
\par
Ultimately, the goal is to implement the spiking TM model on a standalone neuromorphic chip.
In this work, we only investigate how the intrinsic properties of the memristive device affect the learning in the spiking TM.
In a concurrent study, we devise a complete design of the neuromorphic circuit implementing the different components of the spiking TM \citep{Siegel22_memristive} and showed in simulations that the system supports successful prediction performance.   
A future study needs to verify these design principles in physical electronic circuits.


\section{Conclusion}

A prerequisite for the success of analog neuromorphic hardware is the identification of hardware devices that can best implement neuroscientific mechanisms and concepts.
ReRAM devices were suggested as promising synaptic elements due to their scalability and energy efficiency.
In this work, we identified that both analog and binary types of ReRAM device can be used to implement well-known plasticity models from the neuroscientific literature
%
and demonstrated their suitability as synaptic elements in the biologically inspired spiking TM model.
%

\clearpage
\section{Methods}
\label{sec:methods}

In the following summarizing tables (\cref{tab:Model-description}), we provide an overview of the network model, the training protocol, and the simulation details. Parameter values can be found in \cref{tab:Model-parameters}.
See \citep{Bouhadjar22_e1010233} for a detailed description of the model.

\subsection{Model tables}

\begin{table}[!ht]
  \centering
  \renewcommand{\arraystretch}{1.1}
  \small
\begin{tabular}{|@{\hspace*{1mm}}p{3cm}@{}|@{\hspace*{1mm}}p{12cm}|}
\hline 
\multicolumn{2}{|>{\color{white}\columncolor{black}}c|}{\textbf{Summary}}\\
\hline
    \textbf{Populations} &  excitatory neurons ($\Epop$), inhibitory neurons ($\Ipop$), external spike sources ($\Xpop$); $\Epop$ and $\Ipop$ composed of $M$ disjoint subpopulations $\mathcal{M}_k$ and $\Ipop_{k}$ ($k=1,\ldots,M$)\\
\hline 
\textbf{Connectivity} &
\begin{itemize}
    \item sparse random connectivity between excitatory neurons (plastic)
    \item local recurrent connectivity between excitatory and inhibitory neurons (static)
\end{itemize}
\\
\hline
\textbf{Neuron model} & 
\begin{itemize}
\item excitatory neurons: leaky integrate-and-fire (LIF) with nonlinear input integration (dendritic action potentials)      
\item inhibitory neurons: leaky integrate-and-fire (LIF)
\end{itemize}
\\
\hline 
\textbf{Synapse model } & exponential postsynaptic currents (PSCs)  \\
\hline 
\textbf{Plasticity } &  homeostatic spike-timing dependent plasticity in excitatory-to-excitatory connections
\\
\hline 
\end{tabular}
\begin{tabular}{|@{\hspace*{1mm}}p{3cm}@{}|@{\hspace*{1mm}}p{5.95cm}@{}|@{\hspace*{1mm}}p{5.95cm}|}
\hline 
\multicolumn{3}{|>{\color{white}\columncolor{black}}c|}{\textbf{Populations}}\\
\hline
\textbf{Name} & \textbf{Elements} & \textbf{Size}\\
  \hline
  $\mathcal{E}=\cup_{i=k}^M\mathcal{M}_k$ & excitatory (E) neurons  & $N_\exc$\\
  \hline
  $\mathcal{I}=\cup_{i=k}^M\mathcal{I}_k$ & inhibitory (I) neurons & $N_\inh$\\
  \hline
  $\mathcal{M}_k$ & excitatory neurons in subpopulation $k$, \mbox{$\mathcal{M}_k\cap\mathcal{M}_l=\emptyset\ (\forall{}k\ne{}l\in[1,M])$} & $n_\exc$ \\
  \hline 
  $\Ipop_{k}$ & inhibitory neurons in subpopulation $k$, \mbox{$\mathcal{I}_k\cap\mathcal{I}_l=\emptyset\ (\forall{}k\ne{}l\in[1,M])$} & $n_\inh$ \\
  \hline 
  $\Xpop=\{x_1,\ldots,x_M\}$ & external spike sources  & $M$ \\
\hline 
\end{tabular}
\begin{tabular}{|@{\hspace*{1mm}}p{1.85cm}@{}|@{\hspace*{1mm}}p{1.85cm}@{}|@{\hspace*{1mm}}p{11.2cm}|}
\hline 
\multicolumn{3}{|>{\color{white}\columncolor{black}}c|}{\textbf{Connectivity}}\\
\hline 
\textbf{Source population} & \textbf{Target population} & \textbf{Pattern}\\
\hline 
  $\Epop$  & $\Epop$ & random;
                       fixed in-degrees $K_i=K_\EE$, delays $d_{ij}=d_{\EE}$, synaptic time constants $\tau_{ij}=d_{\EE}$ plastic weights $G_{ij}\in\{0,\overline{G}_{ij}\}$
                       ($\forall{}i\in{\Epop},\,\forall{}j\in{\Epop}$; ``$\EE$ connections'') \\
\hline 
  $\mathcal{M}_k$  & $\Ipop_k$ & all-to-all;
                       fixed delays $d_{ij}=d_{\IE}$, synaptic time constants $\tau_{ij}=\tau_{\IE}$, and weights $G_{ij}=G_\IE$
                       ($\forall{}i\in\mathcal{M}_k,\,\forall{}j\in{}\Ipop_k,\,\forall{}k\in[1,M]$; ``$\IE$ connections'') \\
\hline 
  $\Ipop_{k}$ & $\mathcal{M}_k$ & all-to-all;
                       fixed delays $d_{ij}=d_{\EI}$, synaptic time constants $\tau_{ij}=\tau_{\EI}$, and weights $G_{ij}=G_\EI$
                       ($\forall{}i\in\Ipop_k,\,\forall{}j\in{}\mathcal{M}_k,\,\forall{}k\in[1,M]$; ``$\EI$ connections'') \\
\hline 
  $\Ipop_{k}$ & $\Ipop_{k}$ & none ($\forall{}k\in[1,M]$; ``$\II$ connections'') \\
\hline
  $\Xpop_{k}=x_k$ & $\mathcal{M}_k$ & one-to-all;
                       fixed delays $d_{ik}=d_{\EX}$, synaptic time constants $\tau_{ij}=\tau_{\EX}$, and weights $J_{ik}=G_\EX$
                       ($\forall{}i\in\mathcal{M}_k,\,\forall{}k\in[1,M]$; ``$\EX$ connections'')  \\
\hline
\multicolumn{3}{|l|}{no self-connections (``autapses''), no multiple connections (``multapses'') }\\
\hline
  \multicolumn{3}{|l|}{all unmentioned connections
  $\mathcal{M}_k\to\Ipop_l$,
  $\Ipop_k\to\mathcal{M}_l$,
  $\Ipop_k\to\Ipop_l$,
  $\mathcal{X}_k\to\mathcal{M}_l$
  ($\forall{}k\ne{}l$)
  are absent}\\
\hline 
\end{tabular}
\caption{Description of the network model (continued on next page). Parameter values are given in \cref{tab:Model-parameters}.}
\label{tab:Model-description} 
\end{table}
\clearpage
\setcounter{table}{\thetable-1}
\begin{table}[ht!]
  \centering
  \small
\begin{tabular}{|@{\hspace*{1mm}}p{3cm}@{}|@{\hspace*{1mm}}p{12cm}|}
  \hline 
  \multicolumn{2}{|>{\color{white}\columncolor{black}}c|}{\textbf{Neuron and synapse}}\\
  \hline
  \multicolumn{2}{|>{\columncolor{lightgray}}c|}{
  \textbf{Neuron}
  }\\
  \hline
  \textbf{Type} & leaky integrate-and-fire (LIF) dynamics \\
  \hline
  \textbf{Description} & dynamics of membrane potential $V_{i}(t)$ of neuron $i$:                 
    \begin{itemize}
    \item emission of the $k$th spike of neuron $i$ at time $t_{i}^{k}$ if
      \begin{equation}
        V_{i}(t_{i}^{k})\geq\theta_i 
      \end{equation}
      with somatic spike threshold $\theta_i$
      \item reset and refractoriness:
      \begin{equation*}
        V_{i}(t)=\Vreset
        \quad \forall{}k,\ \forall t \in \left(t_{i}^{k},\,t_{i}^{k}+\tau_{\text{ref},i}\right]
      \end{equation*}
      with refractory time $\tau_{\text{ref},i}$ and reset potential $\Vreset$
      \item subthreshold dynamics:
      \begin{equation}
        \label{eq:lif}
        \tau_{\text{m},i}\dot{V}_i(t)=-V_i(t)+R_{\text{m},i} I_i(t)
      \end{equation}
      with membrane resistance $R_{\text{m},i}=\dfrac{\tau_{\text{m},i}}{C_{\text{m},i}}$, membrane time constant $\tau_{\text{m},i}$, and total synaptic input current $I_i(t)$

                 
    \item $\tau_{\text{m},i}=\tau_\text{m,E}$, $C_{\text{m},i}=C_\text{m}$, $\theta_i=\theta_\text{E}$, $\tau_{\text{ref},i}=\tau_\text{ref,E}$ ($\forall i\in\Epop$)
    \item $\tau_{\text{m},i}=\tau_{\text{m},I}$, $C_{\text{m},i}=C_\text{m}$, $\theta_i=\theta_\text{I}$, $\tau_{\text{ref},i}=\tau_\text{ref,I}$ ($\forall i\in\Ipop$)
    
  \end{itemize}\\
  \hline
  \multicolumn{2}{|>{\columncolor{lightgray}}c|}{\textbf{Synapse}}\\
  \hline
  \textbf{Type} & exponential or alpha-shaped postsynaptic currents (PSCs) \\
  \hline
  \textbf{Description} &                 
    \begin{itemize}
      \item total synaptic input current
      \begin{equation}
        \begin{aligned}
          I_i(t) &= I_{\text{ED},i}(t) + I_{\text{EX},i}(t) + I_{\text{EI},i}(t) ,\ \forall i\in\Epop \\
          I_i(t) &= I_{\text{IE},i}(t) ,\ \forall i\in\Ipop
        \end{aligned}
      \end{equation}
      with dendritic, inhibitory, external and excitatory input currents $I_{\text{ED},i}(t)$, $I_{\text{EI},i}(t)$, $I_{\text{EX},i}(t)$,  $I_{\text{IE},i}(t)$ evolving according to
      \begin{equation}
        \label{eq:dendritic_current}
          I_{\text{ED},i}(t)=\sum_{j\in\Epop}(\alpha_{ij}*s_j)(t-d_{ij})
    \end{equation}
    \qquad with $\alpha_{ij}(t)=V_\text{read} G_{ij} \dfrac{e}{\tau_{\text{ED}}} t e^{-t/\tau_{\text{ED}}} \Theta(t)$
    and
    $
    \Theta(t)=
    \begin{cases}1 & t \ge 0 \\ 0 & \text{else} \end{cases}
    $
    \begin{equation}
      \label{eq:EI_current}
      \tau_\text{EI}\dot{I}_{\text{EI},i} = -I_{\text{EI},i}(t) + V_\text{read} \sum_{j\in\Ipop} G_{ij} s_j(t-d_{ij})
    \end{equation}
    \begin{equation}
      \label{eq:EX_current}
      \tau_\text{EX}\dot{I}_{\text{EX},i} = -I_{\text{EX},i}(t) + V_\text{read} \sum_{j\in\Xpop} G_{ij} s_j(t-d_{ij})
    \end{equation}
    \begin{equation}
      \label{eq:IE_current}
      \tau_\text{IE}\dot{I}_{\text{IE},i} = -I_{\text{IE},i}(t) + V_\text{read} \sum_{j\in\Epop} G_{ij} s_j(t-d_{ij})
    \end{equation}
    with $\tau_\text{EX}$, $\tau_\text{EI}$, and $\tau_\text{IE}$ synaptic time constants of EX, EI, and IE connections, respectively, $G_{ij}$ the synaptic weight, and $V$ the read voltage
    \item presynaptic spike trains $s_j(t)=\sum_k \delta(t-t_k^j)$
    \item dAP generation:
      \begin{itemize}
      \item emission of $l$th dAP of neuron $i$ at time $t_i^l$ if $ I_{\text{ED},i}(t_{i}^{l})\geq\theta_{\text{dAP}}$
      \item dAP current plateau:
      \begin{equation}
        \label{eq:dAP_current_nonlinearity}
        I_{\text{ED},i}(t) = I_\text{dAP}
        \quad\forall{}l,\ \forall t \in \left(t_{i}^{l},\,t_{i}^{l}+\tau_\text{dAP}\right]    
    \end{equation}
    with
    dAP current plateau amplitude $I_\text{dAP}$,
    dAP current duration $\tau_\text{dAP}$, and
    dAP activation threshold $\theta_{\text{dAP}}$. 

      \end{itemize}


    \end{itemize} \\
   \hline 
\end{tabular}
\caption{Description of the network model (continued on next page). Parameter values are given in \cref{tab:Model-parameters}.}
\end{table}
\clearpage
\setcounter{table}{\thetable-1}
\begin{table}[ht!]
  \centering
  \small
  \begin{tabular}{|@{\hspace*{1mm}}p{3cm}@{}|@{\hspace*{1mm}}p{12.cm}|}
  \hline 
  \multicolumn{2}{|>{\color{white}\columncolor{black}}c|}{\textbf{Plasticity}}\\
  \hline
  \textbf{Type} & Hebbian-type plasticity and dAP-rate homeostasis \\
  \hline
  \textbf{EE synapses} &
    \begin{itemize}
        \item Hebbian plasticity described in \cref{sec:reram_model}
        \item homeostatic control: 
        \begin{itemize}
            \item if $z_i(t) > z^*$: a depression pulse is applied (see  \cref{eq:analog_synapse} or \cref{eq:binary_synapse})
            \item if $z_i(t) \leq z^*$: a potentiation pulse is applied (see  \cref{eq:analog_synapse} or \cref{eq:binary_synapse})
        \end{itemize}    
            with the dAP trace $z_i(t)$ and target dAP activity $z^*$.
        \item dAP trace $z_i(t)$ of postsynaptic neuron $i$, evolving according to
        \begin{equation*}
          \frac{dz_i}{dt} = -\tau_\text{h}^{-1} z_i(t) + \sum_k \delta(t-t_{\text{dAP},i}^k)
        \end{equation*}
        with onset time $t_{\text{dAP},i}^k$ of the $k$th dAP, homeostasis time constant $\tau_\text{h}$
    \end{itemize}
    \vspace*{1ex}\\ 
  \hline 
  \textbf{all other synapses} & non-plastic
  \\
  \hline
\end{tabular}
\caption{Description of the network model (continued on next page). Parameter values are given in \cref{tab:Model-parameters}.}
\end{table}

\setcounter{table}{\thetable-1}
\begin{table}
  \small
  \begin{tabular}{|@{\hspace*{1mm}}p{15.15cm}|}
  \multicolumn{1}{|>{\color{white}\columncolor{black}}c|}{\textbf{Input}}\\
    \begin{itemize}
    \item repetitive stimulation of the network using the same
      set $\mathcal{S}=\{s_1,\ldots,s_{S}\}$ of
      sequences $s_i=\seq{$\zeta_{i,1}$, $\zeta_{i,2}$,\ldots, $\zeta_{i,C_i}$}$ of
      ordered discrete items $\zeta_{i,j}$ 
      with number of sequences $S$ and length $C_i$ of $i$th sequence
      \item presentation of sequence element $\zeta_{i,j}$ at time $t_{i,j}$ modeled by a single spike $x_k(t)=\delta(t-t_{i,j})$ generated by the corresponding external source $x_k$
      \item inter-stimulus interval $\Delta{}T=t_{i,j+1}-t_{i,j}$ between subsequent sequence elements $\zeta_{i,j}$ and $\zeta_{i,j+1}$ within a sequence $s_i$
      \item inter-sequence time interval $\Delta{}T_\text{seq}=t_{i+1,1}-t_{i,C_i}$ between subsequent sequences $s_i$ and $s_{i+1}$
      \item example sequence sets: 
        \begin{itemize}
        \item sequence set: $\mathcal{S}$=\{\seq{A,D,B,E,I}, \seq{F,D,B,E,C}, \seq{H,L,J,K,D}, \seq{G,L,J,K,E}\}
        \end{itemize}
  \end{itemize}
  \\
  \hline
  \multicolumn{1}{|>{\color{white}\columncolor{black}}c|}{\textbf{Output}} \\
    \hline \\
    \begin{itemize}
    \item somatic spike times $\{t_i^k | \forall{}i\in\mathcal{E},k=1,2,\ldots \}$
    \item dendritic currents $I_{\text{ED},i}(t)$ ($\forall{}i\in\mathcal{E}$)
    \end{itemize}
  \vspace{1ex}
  \end{tabular}
  \begin{tabular}{|@{\hspace*{1mm}}p{15.15cm}|}
  \multicolumn{1}{|>{\color{white}\columncolor{black}}c|}{\textbf{Initial conditions and network realizations}} \\
    \begin{itemize}
    \item membrane potentials: $V_i(0)=V_\text{r}$ ($\forall{}i\in\mathcal{E}\cup\mathcal{I}$)
    \item dendritic currents: $I_{\text{ED},i}(0)=0$ ($\forall{}i\in\mathcal{E}$)
    \item external currents: $I_{\text{EX},i}(0)=0$ ($\forall{}i\in\mathcal{E}$)
    \item inhibitory currents: $I_{\text{EI},i}(0)=0$ ($\forall{}i\in\mathcal{E}$)
    \item excitatory currents: $I_{\text{IE},i}(0)=0$ ($\forall{}i\in\mathcal{I}$)
    \item synaptic permanences: $P_{ij}(0)=P_{\text{min},ij}$ with $P_{\text{min},ij}\sim\mathcal{U}(P_{0,\text{min}},P_{0,\text{max}})$
      ($\forall{}i,j\in\mathcal{E}$)
    \item synaptic weights: $\overline{G}_{ij}(0)=G_{\text{min},ij}$ with $G_{\text{min},ij}\sim\mathcal{U}(G_{0,\text{min}},G_{0,\text{max}})$ ($\forall{}i,j\in\mathcal{E}$) (analog synapse)
    \item synaptic weights: $\overline{G}_{ij}(0)=G_\text{min}$ ($\forall{}i,j\in\mathcal{E}$) (binary synapse)
    \item spike traces: $x_i(0)=0$ ($\forall{}i\in\mathcal{E}$)
    \item dAP traces: $z_i(0)=0$ ($\forall{}i\in\mathcal{E}$)
    \item potential connectivity and initial permanences randomly and independently drawn for each network realization
    \end{itemize}\\
    \hline
  \end{tabular}
  \begin{tabular}{|@{\hspace*{1mm}}p{15.15cm}|}
  \multicolumn{1}{|>{\color{white}\columncolor{black}}c|}{\textbf{Simulation details}}\\
  \hline
  \begin{itemize}
  \item network simulations performed in NEST \citep{Gewaltig_07_11204} version 3.0 \citep{Nest30}
  \item definition of excitatory neuron model using NESTML \citep{Plotnikov16_93, Nagendra_babu_pooja_2021_4740083}
  \item synchronous update using exact integration of system dynamics on discrete-time grid with step size $\dtsim$ \citep{Rotter99a}
  \end{itemize}
  \\
  \hline
\end{tabular}
\caption{Description of the network model. Parameter values are given in \cref{tab:Model-parameters}.}
\end{table}

\clearpage
\subsection{Model and simulation parameters}

\begin{table}[ht!]
  \centering
  \small
\renewcommand{\arraystretch}{1.2}
\begin{tabular}{|@{\hspace*{1mm}}p{3cm}@{}|@{\hspace*{1mm}}p{4cm}@{}|@{\hspace*{1mm}}p{8.1cm}|}
\hline
\textbf{Name} & \textbf{Value} & \textbf{Description}\\
\hline                               
\multicolumn{3}{|>{\columncolor{lightgray}}c|}{\textbf{Network}}\\
\hline 
$N_\exc$ & $1800$ & total number of excitatory neurons \\
\hline
$N_\inh$ & $12$ & total number of inhibitory neurons \\
\hline
$M$ & $14$ & number of excitatory subpopulations (= number of external spike sources)\\
\hline
\textcolor{gray}{$\nE$} & \textcolor{gray}{$N_\exc/M=150$} & number of excitatory neurons per subpopulation \\
\hline
\textcolor{gray}{$\nI$} & \textcolor{gray}{$N_\inh/M=1$} & number of inhibitory neurons per subpopulation \\
\hline
$\rho$ & $20$ & (target) number of active neurons per subpopulation after learning = minimal number of coincident excitatory inputs required to trigger a spike in postsynaptic inhibitory neurons \\
\hline 
\multicolumn{3}{|>{\columncolor{lightgray}}c|}{\textbf{(Potential) Connectivity}}\\
\hline 
$\KEE$ & $450$ & number of excitatory inputs per excitatory neuron ($\EE$ in-degree) \\
\hline 
\textcolor{gray}{$p$} & \textcolor{gray}{$K_{\exc\exc}/N_\exc=0.25$} & probability of potential (excitatory) connections \\
\hline 
\textcolor{gray}{$\KEI$} & \textcolor{gray}{$n_\inh=1$} & number of inhibitory inputs per excitatory neuron ($\EI$ in-degree) \\
\hline 
\textcolor{gray}{$\KIE$} & \textcolor{gray}{$\nE$} & number of excitatory inputs per inhibitory neuron ($\IE$ in-degree) \\
\hline 
$\KII$ & $0$ & number of inhibitory inputs per inhibitory neuron ($\II$ in-degree) \\
\hline 
\multicolumn{3}{|>{\columncolor{lightgray}}c|}{\textbf{Excitatory neurons}}\\
\hline 
$\tau_\text{m,E}$ & $10\ms$ & membrane time constant \\
\hline 
$\tau_\text{ref,E}$ & $20\ms$ & absolute refractory period \\
\hline 
$\CM$ & $250\,\mu{}F$ & membrane capacity \\
\hline 
$\Vreset$ & $0\mV$ & reset potential \\
\hline 
$\theta_\text{E}$ & $30\mV$ & somatic spike threshold \\
\hline 
$ I_\text{dAP}$ & $200\muA$ &  dAP current plateau amplitude\\
\hline 
$\tau_\text{dAP}$ & $60\ms$ & dAP duration\\
\hline 
$\theta_{\text{dAP}}$ & see \cref{eq:fix_dAP} & dAP threshold \\
\hline 
\multicolumn{3}{|>{\columncolor{lightgray}}c|}{\textbf{Inhibitory neurons}}\\
\hline
$\tau_\text{m,I}$ & $5\ms$ & membrane time constant\\
\hline 
$\tau_\text{ref,I}$ & $2\ms$ & absolute refractory period\\
\hline 
$\CM$ & $250\,\mu{}F$ & membrane capacity\\
\hline 
$\Vreset$ & $0\mV$ & reset potential\\
\hline 
$\theta_\text{I}$ & $15\mV$ & spike threshold\\
\hline
\end{tabular}
\caption{Model and simulation parameters (continued on next page). Parameters derived from other parameters are marked in gray.}
\label{tab:Model-parameters} 
\end{table}
\setcounter{table}{\thetable-1}
\begin{table}[ht!]
  \centering
  \small
\begin{tabular}{|@{\hspace*{1mm}}p{3cm}@{}|@{\hspace*{1mm}}p{4cm}@{}|@{\hspace*{1mm}}p{8.1cm}|}
\hline
\textbf{Name} & \textbf{Value } & \textbf{Description}\\
\hline
\multicolumn{3}{|>{\columncolor{lightgray}}c|}{\textbf{Synapse}}\\
\hline
$\tilde{G}_\text{IE}$ & $0.9\mV$  & weight of IE connections (EPSP amplitude) \\
\hline
$\GIE$ & $581.19\muS$ & weight of IE connections (EPSC amplitude) \\
\hline
$\tilde{G}_\text{EI}$ & $-60\mV$ & weight of EI connections (IPSP amplitude) \\
\hline 
$\GEI$ & $-19373.24\muS$ & weight of EI connections (IPSC amplitude) \\
\hline 
$\tilde{G}_\text{EX}$ & $33\mV$ & weight of EX connections (EPSP amplitude) \\
\hline 
$\GEX$ & $6168.31\muS$ & weight of EX connections (EPSC amplitude) \\
\hline 
${\tau}_{\EE}$ & $2 \ms$ & synaptic time constant of EE connections\\
\hline 
${\tau}_{\IE}$ & $0.5 \ms$ & synaptic time constant of IE connections\\
\hline 
${\tau}_{\EI}$ & $1 \ms$ & synaptic time constant of EI connections\\
\hline 
${\tau}_{\EX}$ & $2 \ms$ & synaptic time constant of EX connection\\
\hline
$d_{\EE}$ & $2\ms$ & delay of EE connections (dendritic)\\
\hline
$d_\IE$ & $0.1\ms$ & delay of IE connections\\
\hline
$d_\EI$ & $0.1\ms$ & delay of EI connections\\
\hline 
$d_\EX$ & $0.1\ms$ & delay of EX connections\\
\hline 
\multicolumn{3}{|>{\columncolor{lightgray}}c|}{\textbf{Plasticity}}\\
\hline
$\lambda_{+}$ & $\{0.02,\ldots,\bm{\lplusa},\ldots,0.42\}$ (analog synapse), \newline $\{0.02,\ldots,\bm{\lplusb},\ldots,0.42\}$ (binary synapse) & potentiation learning rate \\
\hline
$\lambda_{-}$ & $\lambda_{+}/\beta$ & depression rate \\
\hline
$\beta$ & $\{0.5,1,2,\bm{3}\}$ & ratio between depression and potentiation learning rates \\
\hline
\textcolor{gray}{$\lambda_\text{h}$} & \textcolor{gray}{$\lambda_{-}$} & homeostasis rate \\
\hline
$\mu_{+}$ & $\{0,\bm{\mplus},1\}$ & weight dependence (potentiation) exponent (default parameter) \\
\hline
$\mu_{-}$ & $\{0,\bm{\mminus},1\}$ & weight dependence (depression) exponent (default parameter) \\
\hline
$\theta_P$ & $10$ & synapse maturity threshold \\
\hline
\textcolor{gray}{$P_{\text{min}, ij}$} & \textcolor{gray}{$\sim\mathcal{U}(P_{0,\text{min}},P_{0,\text{max}})$} & minimum permanence \\
\hline
\textcolor{gray}{$G_{\text{min}, ij}$} & \textcolor{gray}{$\sim\mathcal{U}(G_{0,\text{min}},G_{0,\text{max}})$} & minimum conductance \\
\hline
$G_\text{max}$ & $\{50,\ldots,\bm{300},\ldots,400\}\muS$ & maximum conductance \\
\hline
$G_{0,\text{min}}$ & $7.5\muS$ & minimal initial conductance  \\
\hline
$G_{0,\text{max}}$ & $12.5\muS$ & maximal initial conductance  \\
\hline
$P_{0,\text{max}}$ & $8$ & maximal initial permanence  \\
\hline
$P_{0,\text{min}}$ & $0$ & minimal initial permanence  \\
\hline
$P_{0,\text{max}}$ & $8$ & maximal initial permanence  \\
\hline
$\sigma_\text{r}$ & $\{0,\ldots,\bm{\sr},\ldots,0.1\}$ & read noise  \\
\hline 
$\sigma_\text{w}$ & $\{0,\ldots,\bm{\sw},\ldots,0.25\}$ & write noise \\
\hline     
$ z^* $ & $1.8$ & target dAP activity \\
\hline
$ \tau_\text{h} $ & $1040\,\text{ms}$  & homeostasis time constant \\
\hline    
\multicolumn{3}{|>{\columncolor{lightgray}}c|}{\textbf{Input}}\\
$S$ & $4$ & number of sequences per set \\
\hline 
$C$ & $5$ & number of characters per sequence \\
\hline
$A$ & $12$ & alphabet length\\
\hline
$\Delta{}T$ & $40\ms$ & inter-stimulus interval \\
\hline
$\Delta{}T_\text{seq}$ & $100\ms$ & inter-sequence interval \\
\hline
\multicolumn{3}{|>{\columncolor{lightgray}}c|}{\textbf{Simulation}}\\
\hline
$\dtsim$ & $0.1\ms$ & time resolution \\
\hline
$K$ & $400$ & number of training episodes \\
\hline 
\end{tabular}
\caption{Model and simulation parameters. Parameters derived from other parameters are marked in gray. Bold numbers depict default values. Curly brackets
depict a set of values corresponding to different experiments.}
\end{table}

\clearpage
\section*{Acknowledgments}
This project was funded by the Helmholtz Association Initiative and Networking Fund (project number SO-092, Advanced Computing Architectures), and the European Union's Horizon 2020 Framework Programme for Research and Innovation under the Specific Grant Agreement No.~785907 (Human Brain Project SGA2) and No.~945539 (Human Brain Project SGA3).
Open access publication was funded by the Deutsche Forschungsgemeinschaft (DFG, German Research Foundation; grant 491111487).

\section*{Author contributions}
All authors conceived and designed the work. 
YB performed the simulations and analyzed and visualized the data. 
YB wrote the first draft of the manuscript.
All authors reviewed the manuscript and approved it for publication. 
YB was supervised by TT and DJW.



\clearpage
\appendixpageoff
\appendixtitleoff
\renewcommand{\appendixtocname}{Supporting information}
\begin{appendices}

\section{Supporting information}
\label{sec:supplemental_materials}

\setcounter{figure}{0}
\renewcommand{\thefigure}{S\arabic{figure}}%
\setcounter{section}{0}
\renewcommand{\thesection}{S}%
\setcounter{table}{0}
\renewcommand{\thetable}{S\arabic{table}}%

\begin{figure}[!h]
  \centering
  \includegraphics{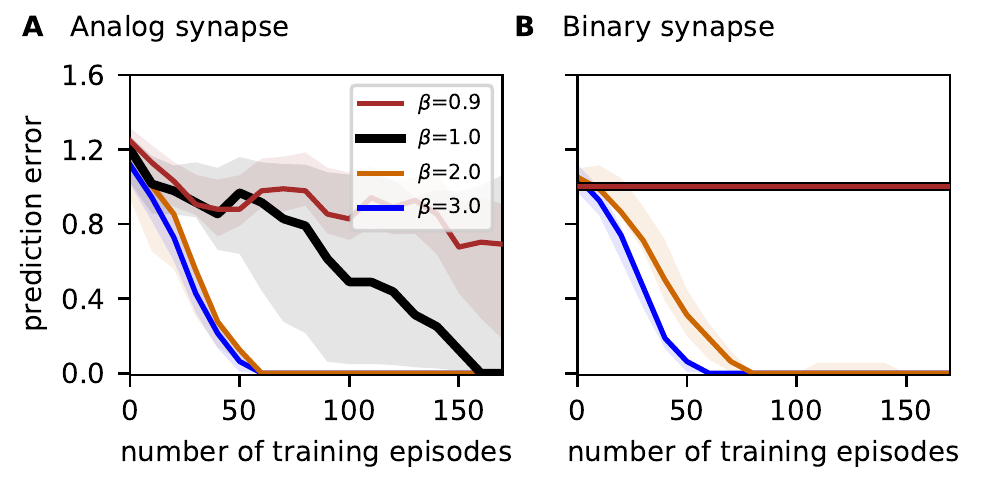}
  \caption{%
    \textbf{Effect of the asymmetry in the learning rates on the prediction performance.}
    Dependence of the prediction error on the number of training episodes for different ratios ($\beta$) between the depression and potentiation learning rates ($\lambda_{-}= \lambda_{+} / \beta$), shown for the network with either analog ($\panellabel{A}$) or binary ($\panellabel{B}$) synapses. 
    The potentiation learning rate $\lambda_{+}$ is fixed to $\lplusa$ for the analog synapse and to $\lplusb$ for the binary synapse.
    Curves and error bands indicate the median as well as the $5\%$ and $95\%$ percentiles across an ensemble of $5$ different network realizations, respectively.
    For remaining parameters see \cref{tab:Model-parameters}.
  }
\label{fig:supp_asy_prediction_error} 
\end{figure}


\end{appendices}

\end{document}